\documentclass{article}
\usepackage{iclr2023_conference,times}

\usepackage{amsmath,amsfonts,bm}

\def\eqref#1{equation~\ref{#1}}

\def\1{\bm{1}}

\DeclareMathAlphabet{\mathsfit}{\encodingdefault}{\sfdefault}{m}{sl}
\SetMathAlphabet{\mathsfit}{bold}{\encodingdefault}{\sfdefault}{bx}{n}

\usepackage[colorlinks, linkcolor=red, anchorcolor=red, citecolor=cyan, breaklinks]{hyperref}
\usepackage{url}

\usepackage{times}
\usepackage{epsfig}
\usepackage{amsmath}
\usepackage{amssymb}
\usepackage{wrapfig}
\usepackage{enumitem}
\usepackage{graphicx}
\usepackage{subfiles}
\usepackage{multirow}
\usepackage{booktabs}
\usepackage{subfigure}

\usepackage{color}
\usepackage{colortbl}
\setlist[itemize]{leftmargin=2em}
\newtheorem{remark}{Guidance}
\newtheorem{proof}{Proof}
\newtheorem{theorem}{Proposition}
\newtheorem{definition}{Theorem}

\usepackage{algorithm}
\usepackage{algpseudocode}

\title{Automated Graph Self-supervised Learning via Multi-teacher Knowledge Distillation}

\iclrfinalcopy

\author{
  Lirong Wu, Yufei Huang, Haitao Lin, Zicheng Liu, Tianyu Fan, and Stan Z. Li \\
  AI Lab, School of Engineering, Westlake University \\
  \texttt{\{wulirong,huangyufei,linhaitao,stan.zq.li\}@westlake.edu.cn} \\
}

\begin{document}

\maketitle

\vspace{-1em}
\begin{abstract}
Self-supervised learning on graphs has recently achieved remarkable success in graph representation learning. With hundreds of self-supervised pretext tasks proposed over the past few years, the research community has greatly developed, and the key is no longer to design more powerful but complex pretext tasks, but to make more effective use of those already on hand. This paper studies the problem of how to automatically, adaptively, and dynamically learn instance-level self-supervised learning strategies for each node from a given pool of pretext tasks. In this paper, we propose a novel multi-teacher knowledge distillation framework for \underline{A}utomated \underline{G}raph \underline{S}elf-\underline{S}upervised \underline{L}earning (\texttt{AGSSL}), which consists of two main branches: \textit{(i) Knowledge Extraction:} training multiple teachers with different pretext tasks, so as to extract different levels of knowledge with different inductive biases; \textit{(ii) Knowledge Integration:} integrating different levels of knowledge and distilling them into the student model. Without simply treating different teachers as equally important, we provide a provable theoretical guideline for how to integrate the knowledge of different teachers, i.e., the integrated teacher probability should be close to the true Bayesian class-probability. To approach the theoretical optimum in practice, two adaptive knowledge integration strategies are proposed to construct a relatively ``good" integrated teacher. Extensive experiments on eight datasets show that \texttt{AGSSL} can benefit from multiple pretext tasks, outperforming the corresponding individual tasks; by combining a few simple but classical pretext tasks, the resulting performance is comparable to other leading counterparts.
\end{abstract}

\vspace{-1em}
\section{Introduction}
\vspace{-0.5em}
Deep learning on graphs \citep{hamilton2017inductive,kipf2016semi,velivckovic2017graph,wu2020comprehensive} has recently achieved remarkable success on a variety of tasks, while such success relies heavily on the massive and carefully labeled data. However, precise annotations are usually very expensive and time-consuming. Recent advances in graph \textit{Self-supervised Learning} (SSL) \citep{wu2021self,xie2021self,liu2021graph} have provided novel insights into reducing the dependency on annotated labels and enable the training on massive unlabeled data. The primary goal of graph SSL is to provide self-supervision for learning transferable knowledge from abundant unlabeled data, through well-designed pretext tasks (in the form of loss functions). There have been hundreds of pretext tasks proposed in the past few years \citep{sun2019infograph,hu2019strategies,xia2022simgrace,xia2021debiased,zhu2020deep,you2020graph,zhang2020iterative}, and different pretext tasks extract different levels of graph knowledge based on different inductive biases. For example, \texttt{PAIRDIS} \citep{jin2020self} captures the inter-node long-range dependencies by predicting the shortest path lengths between nodes, while \texttt{PAR} \citep{you2020does} extracts topological information by predicting the graph partitions of nodes. With so many ready-to-use pretext tasks already on hand, as opposed to designing more complex pretext tasks (for minor performance gains), a more promising problem is \textit{how to automatically, adaptively, and dynamically leverage multiple existing pretext tasks effectively}.

\begin{figure*}[!tbp]
    \vspace{-2em}
	\begin{center}
		\subfigure[Dataset-level Dependency ]{\includegraphics[width=0.28\linewidth]{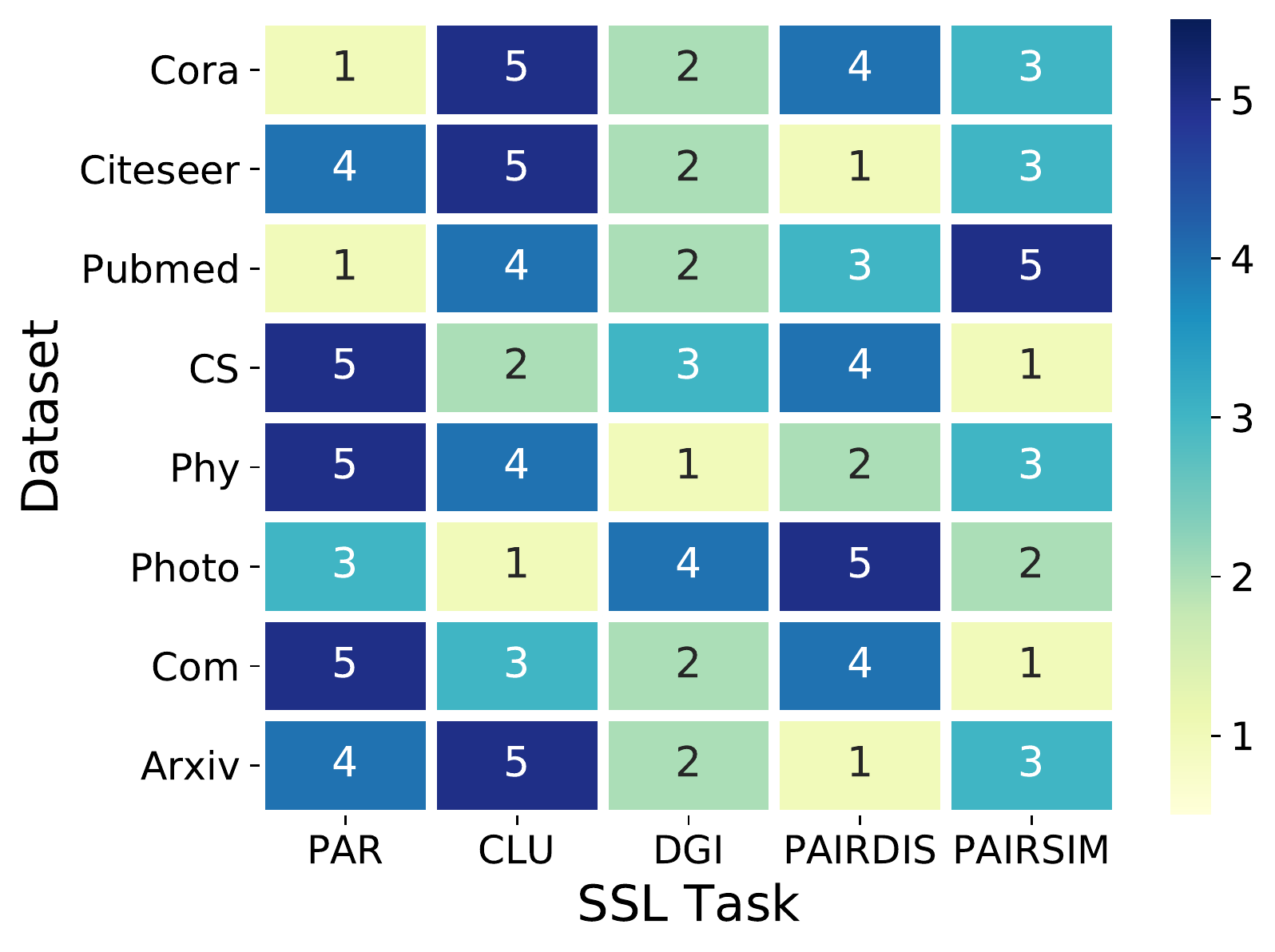} \label{fig:1a}}
		\subfigure[Task-level Compatibility]{\includegraphics[width=0.29\linewidth]{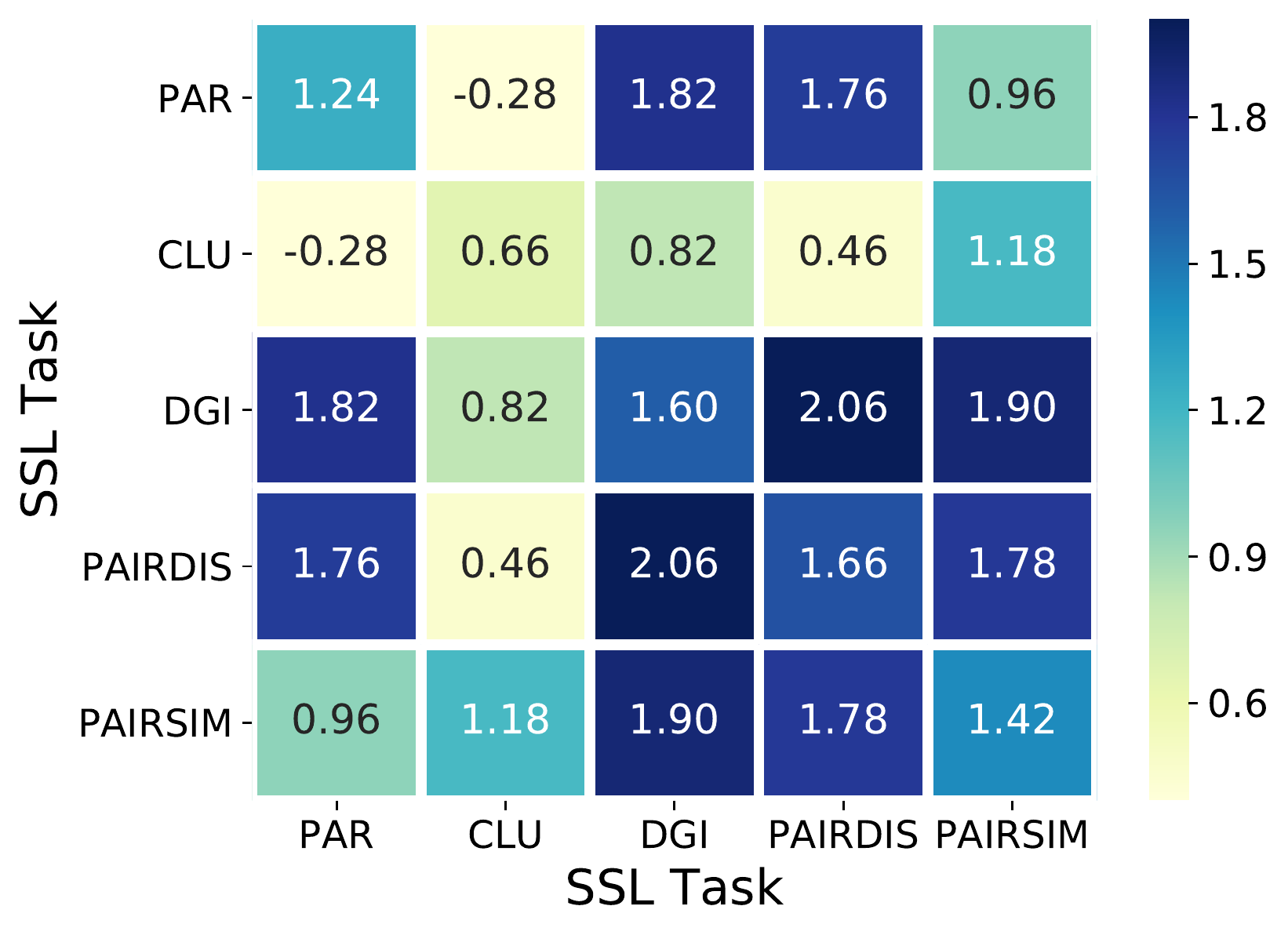} \label{fig:1b}}
		\subfigure[Node-level Localization]{\includegraphics[width=0.39\linewidth]{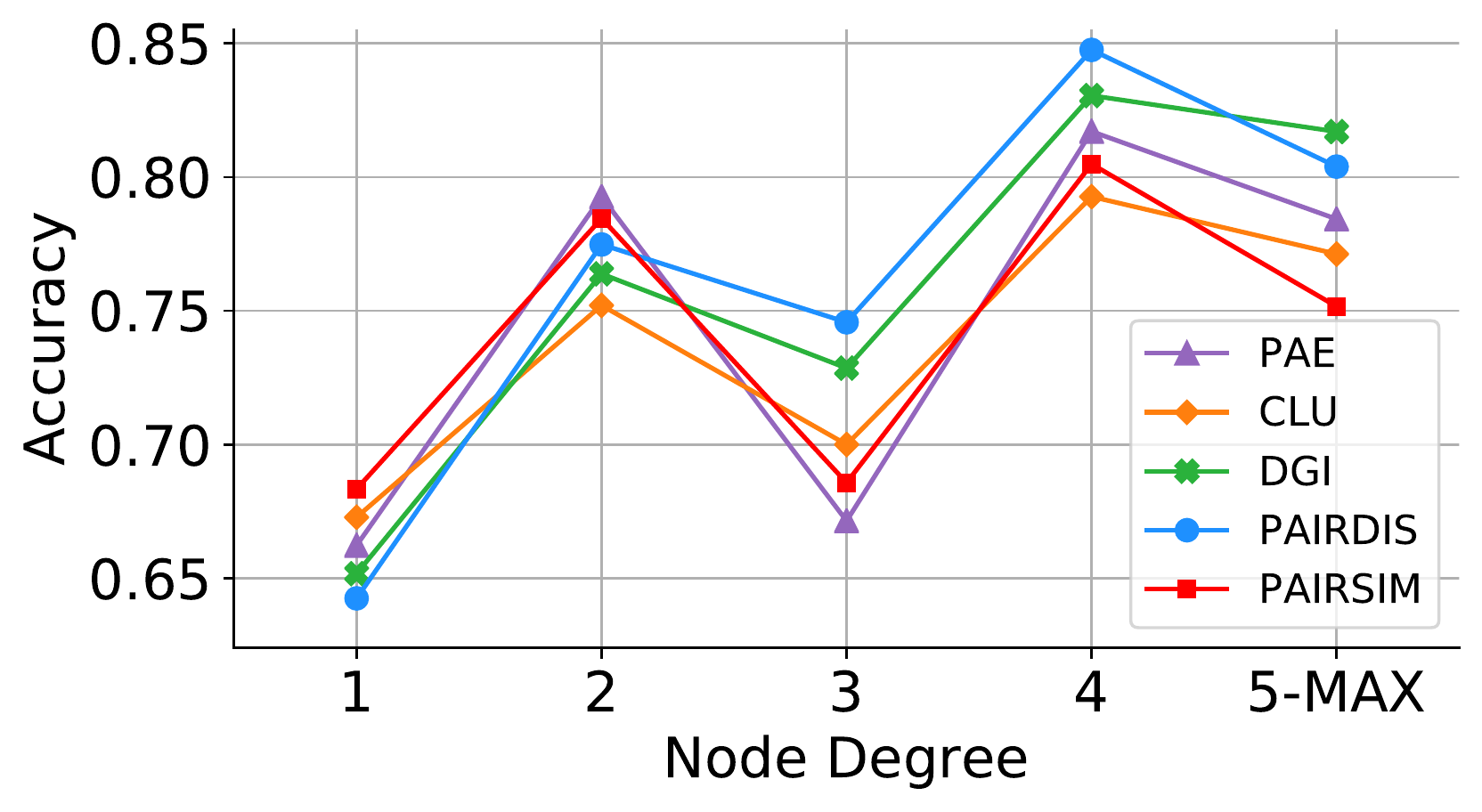} \label{fig:1c}}
	\end{center}
	\vspace{-1em}
	\caption{\textbf{(a)} Performance ranking (1: best, 5: poorest) of different pretext tasks on eight datasets. \textbf{(b)} Performance gains or drops over the vanilla GCNs on \texttt{Citeseer} when combining two tasks. \textbf{(c)} Classification accuracy of nodes with different node degrees across five pretext tasks on \texttt{Citeseer}.}
	\vspace{-0.5em}
	\label{fig:1}
\end{figure*}

The investigations on five self-supervised pretext tasks, including \texttt{PAR}, \texttt{CLU} \citep{you2020does}, \texttt{DGI} \citep{velickovic2019deep}, \texttt{PAIRDIS} and \texttt{PAIRSIM} \citep{jin2020self}, yielded some interesting observations, from which we identify three main challenges for automated graph SSL: \textit{(i) Dataset-level dependency.} The node classification performance rankings in Fig.~\ref{fig:1a} indicate that different pretext tasks have distinct downstream performance across eight datasets, and the success of pretext tasks strongly depends on the dataset characteristics. \textit{(ii) Task-level compatibility.} The results reported in Fig.~\ref{fig:1b} are quite counterintuitive, where learning representations with multiple pretext tasks does not necessarily lead to better performance, because pretext tasks involving different inductive biases may be incompatible with each other. \textit{(iii) Node-level Localization.} We present the classification accuracy of nodes with different node degrees across five pretext tasks in Fig.~\ref{fig:1c}, from which we observe that different nodes may require localized pretext tasks; for example, high-degree nodes benefit more from \texttt{DGI} and \texttt{PAIRDIS}, while low-degree nodes prefer \texttt{CLU} and \texttt{PAIRSIM}.

Given an unseen graph dataset and a pool of pre-given pretext tasks, this paper aims to design an automated graph self-supervised learning framework that meets the following four desirable properties: \textit{(i)} \textbf{automated}, completely free of human labor or trial-and-error on pretext task choices; \textit{(ii)} \textbf{adaptive}, generalizing well to various graph datasets; \textit{(iii)} \textbf{dynamic}, allowing for different types of pretext tasks at different training stages; \textit{(iv)} \textbf{localized}, learning a customized combination of pretext tasks for each node separately and providing instance-level explanations on the importance of different tasks for each node. Compared to previous hand-crafted, dataset-specific task selection, this paper achieves an unprecedented degree of flexibility. We summarize our contributions as

\begin{itemize}
    \item  We propose a novel multi-teacher knowledge distillation framework for \underline{A}utomated \underline{G}raph \underline{S}elf-\underline{S}upervised \underline{L}earning (\texttt{AGSSL}), which can automatically, adaptively, and dynamically learn instance-level self-supervised learning strategies for each node separately.
    \item We provide a provable theoretical guideline for how to integrate the knowledge of different teachers, i.e., the integrated teacher probability should be close to the true class-Bayesian probability. Moreover, we prove theoretically that the optimal integrated teacher probability can monotonically approach the Bayesian class-probability as the number of teachers increases.
    \item Two knowledge integration strategies are proposed to construct a relatively ``good" teacher by adaptively adjusting the weights of different knowledge for each node during distillation.
    \item Extensive experiments on eight datasets demonstrate that \texttt{AGSSL} can benefit from multiple pretext tasks and significantly improve the performance of individual tasks. By combining only a few simple pretext tasks, the resulting performance is comparable to other leading counterparts.
\end{itemize}

\vspace{-1em}
\section{Preliminaries} \label{sec:2}
\vspace{-0.5em}
\textbf{Notations.} Let $\mathcal{G}=(\mathcal{V}, \mathcal{E}, \mathbf{X})$ denote an attributed graph, where $\mathcal{V}$ is the set of $|\mathcal{V}|=N$ nodes with features $\mathbf{X}=\left[\mathbf{x}_{1}, \mathbf{x}_{2}, \cdots, \mathbf{x}_{N}\right]\in \mathbb{R}^{N \times d}$ and $\mathcal{E} \subseteq \mathcal{V} \times \mathcal{V}$ is the set of $|\mathcal{E}|$ edges between nodes. Following the common semi-supervised node classification setting, only a subset of node $\mathcal{V}_L=\{v_1,v_2,\cdots,v_L\}$ with corresponding labels $\mathcal{Y}_L=\{y_1,y_2,\cdots,y_L\}$ are known, and we denote the labeled set as $\mathcal{D}_L=(\mathcal{V}_L,\mathcal{Y}_L)$ and unlabeled set as $\mathcal{D}_U=(\mathcal{V}_U,\mathcal{Y}_U)$, where $\mathcal{V}_U=\mathcal{V} \backslash \mathcal{V}_L$. The task of node classification aims to learn a GNN encoder $f_{\theta}(\cdot)$ and a linear prediction head $g_{\omega}(\cdot)$ with the task loss $\mathcal{L}_{\mathrm{task}}(\theta,\omega)$ on labeled data $\mathcal{D}_L$, so that they can be used to infer the labels $\mathcal{Y}_U$.

\begin{figure*}[!tbp]
    \vspace{-2em}
	\begin{center}
		\subfigure[Joint Training (\texttt{JT})]{\includegraphics[width=0.5\linewidth]{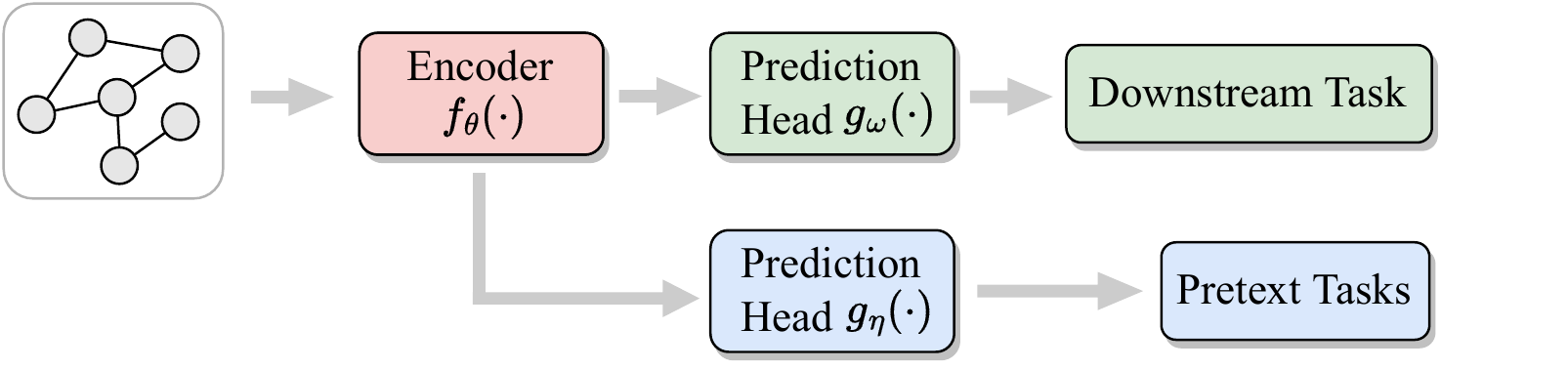}} 
		\subfigure[Pre-train\&Fine-tune (\texttt{P\&F})]{\includegraphics[width=0.46\linewidth]{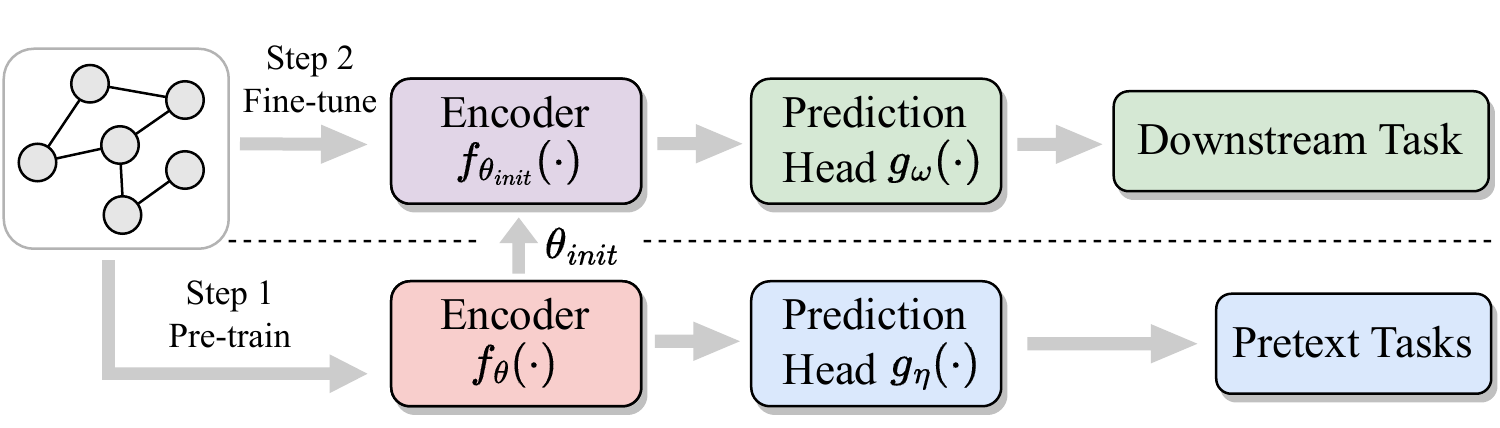}}
	\end{center}
	\vspace{-1em}
	\caption{Illustration of the two training strategies, namely Joint Training and Pre-train\&Fine-tune.}
	\vspace{-0.5em}
	\label{fig:2}
\end{figure*}

\textbf{Problem Statement.} Given a GNN encoder $f_{\theta}(\cdot)$, a linear prediction head $g_\omega(\cdot)$, and a set of $K$ self-supervised losses $\{\mathcal{L}_{\mathrm{ssl}}^{(1)}(\theta,\eta_1),\mathcal{L}_{\mathrm{ssl}}^{(2)}(\theta,\eta_2),\cdots,\mathcal{L}_{\mathrm{ssl}}^{(K)}(\theta,\eta_K)\}$ with prediction heads $\{g_{\eta_k}(\cdot)\}_{k=1}^K$, two common strategies for combining self-supervised losses $\{\mathcal{L}_{\mathrm{ssl}}^{(k)}(\theta,\eta_k)\}_{k=1}^K$ and semi-supervised loss $\mathcal{L}_{\mathrm{task}}(\theta,\omega)$ are \textit{Joint Training} (\texttt{JT}) and \textit{Pre-train\&Fine-tune} (\texttt{P\&F}), as shown in Fig.~\ref{fig:2}. The \textit{Joint Training} strategy jointly train the entire model under the supervision of downstream task and pretext tasks, which can be considered as a kind of multi-task learning, defined as
\vspace{-0.3em}
\begin{equation}
\theta^*, \omega^*, \{\eta_k^*\}_{k=1}^K=\mathop{\arg\min}_{\theta, \omega, \{\eta_k\}_{k=1}^K} \mathcal{L}_{\mathrm{task}}(\theta, \omega)+\alpha \sum_{k=1}^K \lambda_k \mathcal{L}_{\mathrm{ssl}}^{(k)}(\theta,\eta_k)
\end{equation}
where $\alpha$ is a trade-off hyperparameter and $\{\lambda_k\}_{k=1}^K$ are task weights. The \textit{Pre-train\&Fine-tune} strategy works in a two-stage manner: (1) Pre-training the GNN encoder $f_\theta(\cdot)$ with self-supervised pretext tasks; and (2) Fine-tuning the pre-trained GNN encoder $f_{\theta_{init}}(\cdot)$ with a prediction head $g_\omega(\cdot)$ under the supervision of a specific downstream task. The learning objective can be formulated as
\begin{equation}
\theta^{*}, \omega^{*}=\arg \min _{(\theta, \omega)} \mathcal{L}_{\mathrm{task}}(\theta_{init}, \omega), \quad \text{s.t.} \text{ } \text{ } \theta_{init}, \{\eta_k^*\}_{k=1}^K=\mathop{\arg\min}_{\theta, \{\eta_k\}_{k=1}^K} \sum_{k=1}^K \lambda_k \mathcal{L}_{\mathrm{ssl}}^{(k)}(\theta,\eta_k)
\end{equation}
A high-level overview of the two strategies is shown in Fig.~\ref{fig:2}. Without loss of generality, we mainly introduce our framework for the \texttt{JT} strategy, leaving extensions to the \texttt{P\&F} strategy in \textbf{Appendix A}.

\textbf{Graph Self-supervised Learning Automation.} A Vanilla solution to combine multiple self-supervised pretext tasks is to set the task weight $\lambda_k=\frac{1}{K}$ ($1 \leq k \leq K$), i.e., to treat different tasks as equally important, but this completely conflicts with the observations from Fig.~\ref{fig:1}. Furthermore, the experimental results in Table.~\ref{tab:1} also show that the performance of simply averaging over different tasks may not only fail to match the performance of individual tasks, but may even be inferior to the vanilla implementation without any pretext tasks. Different from hand-crafted task weight settings, AutoSSL \citep{jin2021automated} proposes to learn a set of task weights $\{\lambda_k\}_{k=1}^K$ such that $f_\theta(\cdot)$ trained with the weighted loss $\sum_{k=1}^K \lambda_k \mathcal{L}_{\mathrm{ssl}}^{(k)}(\theta,\eta_k)$ can extract meaningful representations. Specifically, AutoSSL formulates the automated self-supervised task search as a bi-level optimization problem and solves it via meta-gradient descent \citep{finn2017model,zugner2019adversarial}, as follows
\begin{equation}
\min_{\{\lambda_k\}_{k=1}^K} \mathcal{H}\big(f_{\theta^*}(\mathcal{G})\big), \text{ }\text{ } \text{s.t.} \text{ } \text{ } \theta^*, \omega^*, \{\eta_k^*\}_{k=1}^K=\mathop{\arg\min}_{\theta, \omega, \{\eta_k\}_{k=1}^K} \mathcal{L}_{\mathrm{task}}(\theta, \omega)+\alpha \sum_{k=1}^K \lambda_k \mathcal{L}_{\mathrm{ssl}}^{(k)}(\theta,\eta_k)
\end{equation}
where $\mathcal{H}(\cdot)$ denotes the quality measure of the node representations $f_{\theta^*}(\mathcal{G})$, and it can be any metric that evaluates the downstream performance, such as the cross-entropy loss on the labeled data $\mathcal{V}_L$ for the semi-supervised learning setting. In practice, how to define $\mathcal{H}(\cdot)$ is usually a heuristic problem that requires a lot of manual labor, which may not satisfy the ``automated" condition. Secondly, adjusting task weights can alleviate, but not completely solve, the task-level compatibility problem, because the loss of a task with small weight can still couple with and affect other tasks \citep{kendall2018multi,chen2018gradnorm}. Thirdly, the learned task weights may fail to explain the importance of different tasks, since the weights of different self-supervised losses can be of different orders of magnitude, and thus high-weighted losses do not necessarily dominate in optimization. Last but not least, AutoSSL only learns \textbf{global} task weights for each dataset, but completely ignores the node-level localization (as found in Fig.~\ref{fig:1c}). Considering these four limitations, we propose a novel multi-teacher knowledge distillation framework for \underline{A}utomated \underline{G}raph \underline{S}elf-\underline{S}upervised \underline{L}earning (\texttt{AGSSL}), which can learn instance-level self-supervised strategies for each node separately.

\begin{figure*}[!tbp]
    \vspace{-2em}
	\begin{center}
		\subfigure[Joint Training]{\includegraphics[width=0.225\linewidth]{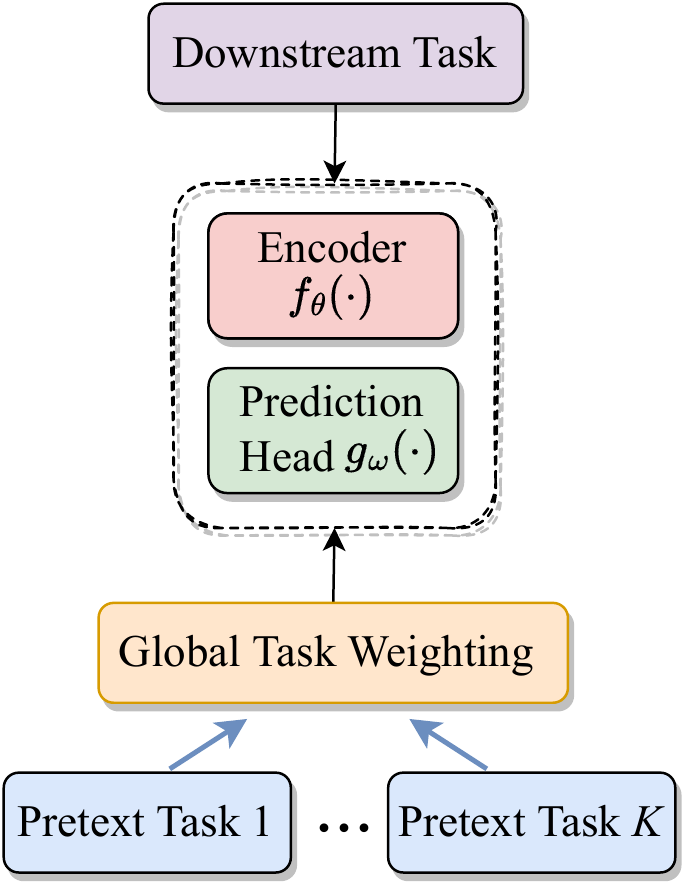}} 
		\subfigure[Multi-teacher Knowledge Distillation for Automated Graph SSL]{\includegraphics[width=0.75\linewidth]{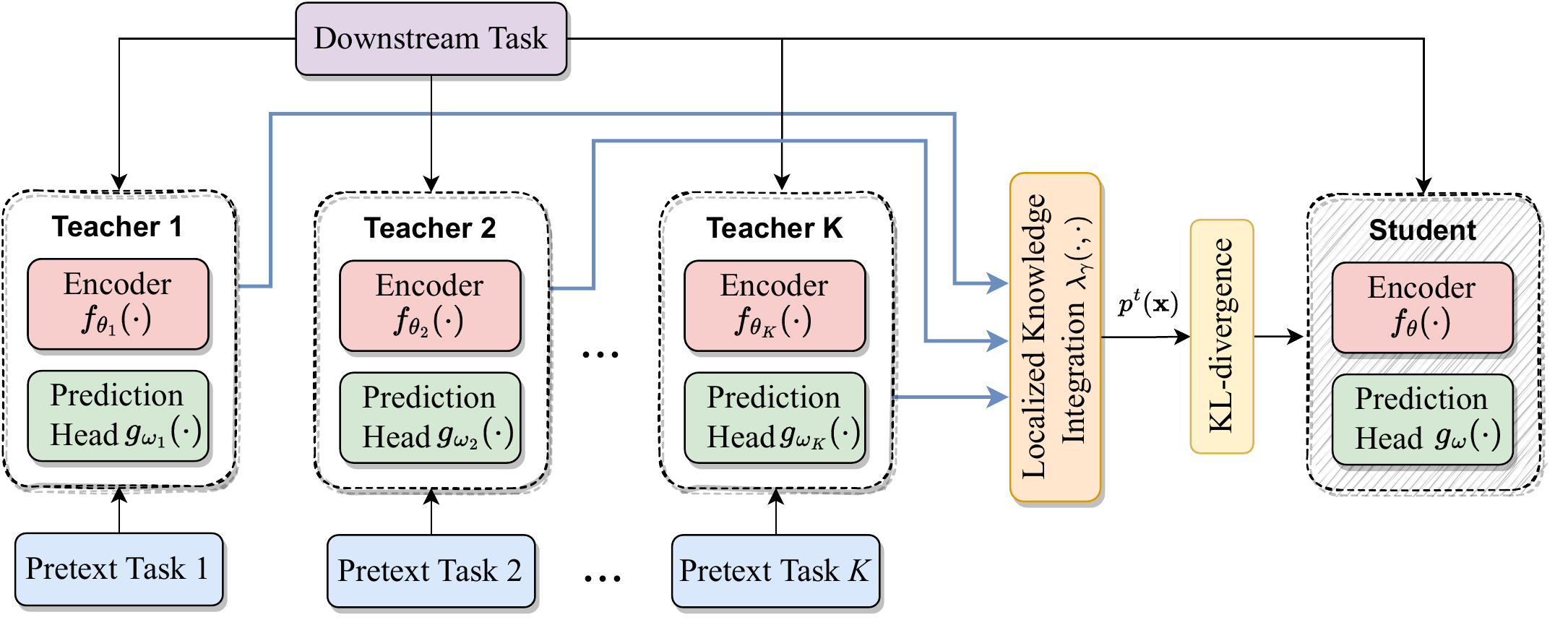}}
	\end{center}
	\vspace{-1em}
	\caption{\textbf{(a)} Conventional multi-task self-supervised learning where the model is jointly trained with multiple (globally) weighted pretext tasks. \textbf{(b)} Proposed multi-teacher knowledge distillation framework, where we train each teacher separately with one pretext task and then apply a localized integration module to integrate different levels of knowledge and distill them into the student model.}
	\vspace{-1em}
	\label{fig:3}
\end{figure*}

\vspace{-0.5em}
\section{Methodology}
\vspace{-0.5em}
\subsection{Multi-teacher Knowledge Distillation} \label{sec:3.1}
Intuitively, training with multiple pretext tasks enables the model to access richer information, which is beneficial for improving performance. However, this holds true only if we can well handle the compatibility problem between pretext tasks. Previous approaches, such as AutoSSL, have attempted to jointly train multiple pretext tasks by adaptively learning global task weights, but this learning process is a black box in which multiple tasks are coupled with each other to produce unexplainable ``mixed" representations. To make matters worse, this renders the training of a primary task vulnerable to other relatively unrelated tasks, leading to suboptimal results. In this paper, we propose a novel multi-teacher knowledge distillation framework - \texttt{AGSSL} in Fig.~\ref{fig:3}, where we train multiple teachers with different pretext tasks to extract different levels of knowledge, which are then integrated through a localized knowledge integration module $\lambda_\gamma(\cdot,\cdot)$ and finally distilled into the student model. The learning objective of \texttt{AGSSL} can be formulated as follows
\vspace{-0.5em}
\begin{equation}
\begin{aligned}
& \min_{\theta, \omega, \gamma} \mathcal{L}_{\mathrm{task}}\big(\theta, \omega\big)+\beta \frac{\tau^2}{N}\sum_{i=1}^N \mathcal{L}_{KL} \Big(\operatorname{softmax}(\mathbf{z}_i/\tau), \sum_{k=1}^K \lambda_\gamma(k,i) \operatorname{softmax}(\mathbf{h}_i^{(k)}/\tau)\Big) \\
& \text{ } \text{s.t.} \text{ } \text{ } \theta_k^*, \omega_k^*, \eta_k^* = \mathop{\arg\min}_{\text{ }\text{ }\text{ }\text{ }\text{ }\text{ }\theta_k, \omega_k, \eta_k} \mathcal{L}_{\mathrm{task}}(\theta_k, \omega_k)+\alpha \mathcal{L}_{\mathrm{ssl}}^{(k)}(\theta_k,\eta_k), \text{ } \text{where} \text{ } \text{ } 1 \leq k \leq K
\end{aligned}
\label{equ:4}
\end{equation}
where $\mathbf{z}_i \!=\! g_\omega(f_\theta(\mathcal{G}, i))$ and $\mathbf{h}_i^{(k)} \!=\! g_{\omega_k^*}(f_{\theta_k^*}(\mathcal{G}, i))$ are the logits of node $v_i$ in the student model and $k$-th teacher model, respectively. Besides, $\mathcal{L}_{KL}(\cdot,\cdot)$ denotes the KL-divergence loss, and $\lambda_\gamma(k,i)$ is a weighting function that outputs the importance weight of $k$-th pretext task for node $v_i$, which satisfies $\sum_{k=1}^K\lambda_\gamma(k,i)=1$. In addition, $\beta$ is a trade-off hyperparameter, $\tau$ is the distillation temperature, and $\tau^2$ is used to keep the gradient stability of this loss. The knowledge distillation is first introduced in \citep{hinton2015distilling}, where knowledge is transferred from a cumbersome teacher to a lightweight student. In this paper, not to get a simpler student model, we adopt the multi-teacher knowledge distillation framework to \textbf{automatically}, \textbf{adaptively}, and \textbf{dynamically} distill different levels of knowledge into one student model. Besides, \texttt{AGSSL} takes full account of the node-level \textbf{localization} and learns a customized knowledge integration strategy for each node through a parameterized function $\lambda_\gamma(\cdot,\cdot)$. Moreover, since the knowledge extracted by different tasks is mapped into the same probability space and the node-level weights satisfy $\sum_{k=1}^K\lambda_\gamma(k,i)\!=\!1$, the weight $\lambda_\gamma(k,i)$ can more truly \textbf{explain} the importance of $k$-th pretext task to node $v_i$.

\vspace{-0.3em}
\subsection{Theoretical Guideline for How to Integrate}
\vspace{-0.2em}
One more problem left to be solved in Sec.~\ref{sec:3.1} is how to integrate different levels of knowledge, that is, how to design and learn a suitable knowledge weighting function $\lambda_\gamma(k,i)$. In this section, we (P1) establish a provable theoretical guideline that tells us how to integrate, i.e., \textit{what is the criteria for constructing a relatively ``good" integrated teacher}; (P2) a theory-guided practical implementation; and (P3) a theoretical justification for the advantages of \texttt{AGSSL} under the multi-task learning setting.

\vspace{0.5em}
\textbf{\emph{(P1)}} Let us define $R(\theta,\omega)\!=\!\mathbb{E}_{\mathbf{x}}\big[\mathbb{E}_{y  \mid \mathbf{x}}\big[\ell\big(y, \operatorname{softmax}(g_\omega(f_\theta(\mathbf{x}))/\tau)\big)\big]\big]\!=\!\mathbb{E}_{\mathbf{x}}\left[\mathbf{p}^{*}(\mathbf{x})^{\top} \mathbf{l}\big(g_\omega(f_\theta(\mathbf{x}))\big)\right]$ as \textit{\textbf{Bayesian objective}}, where $\mathbf{p}^{*}(x) \doteq[\mathbb{P}(y | x)]_{y \in[C]}$ is defined as the Bayesian probability. $\mathbf{l}\big(g_\omega(f_\theta(\mathbf{x}))\big)\!=\!\big(\ell(1, \operatorname{softmax}(g_\omega(f_\theta(\mathbf{x}))/\tau)),\cdots,\ell(C, \operatorname{softmax}(g_\omega(f_\theta(\mathbf{x}))/\tau))\big)$ is the loss vector, where $\ell(\cdot,\cdot)$ is the cross-entropy loss and $C$ is the number of category. To simplify the notations, we can set $\widetilde{\mathbf{z}}_i\!\doteq\!\operatorname{softmax}(\mathbf{z}_i/\tau)$, $\widetilde{\mathbf{h}}_i^{(k)}\!\doteq\!\operatorname{softmax}(\mathbf{h}_i^{(k)}/\tau)$, and $\mathbf{p}^{\mathrm{t}}(\mathbf{x}_i)\!\doteq\!\sum_{k=1}^K \lambda_\gamma(k,i)\widetilde{\mathbf{h}}_i^{(k)}$, and then we can rewrite the the second distillation term of Eq.~(\ref{equ:4}) as the \textit{\textbf{distillation objective}}, as follows
\begin{equation}
    \frac{1}{N}\sum_{i=1}^N \mathcal{L}_{KL} \Big(\widetilde{\mathbf{z}}_i, \sum_{k=1}^K \lambda_\gamma(k,i) \widetilde{\mathbf{h}}_i^{(k)}\Big) \propto \frac{1}{N}\sum_{i=1}^N \mathbf{p}^{\mathrm{t}}(\mathbf{x}_i)\mathbf{l}\big(g_\omega(f_\theta(\mathbf{x}_i))\big)\doteq \widetilde{R}(\theta,\omega)
    \label{equ:5}
\end{equation}
where the detailed derivation of Eq.~(\ref{equ:5}) is available in \textbf{Appendix B}. Previous work \citep{menon2021statistical} has provided a statistical perspective on single-teacher knowledge distillation, where a Bayesian teacher providing true class probabilities $\{\mathbf{p}^*(\mathbf{x}_i)\}_{i=1}^N$ can lower the variance of the \textit{\textbf{downstream objective}} $\mathcal{L}_{\mathrm{task}}(\theta,\omega)=\frac{1}{N}\sum_{i=1}^N \mathbf{e}^T_{y_i}\mathbf{l}\big(g_\omega(f_\theta(\mathbf{x}_i))\big)$, where $\mathbf{e}^T_{y_i}$ is the one-hot label of node $v_i$; the reward of reducing variance is beneficial for improving generalization \citep{maurer2009empirical}. However, the teacher probabilities $\{\widetilde{\mathbf{h}}_i^{(k)}\}_{k=1}^K$ and Bayesian probability $\mathbf{p}^*(\mathbf{x}_i)$ are very likely to be \textit{linearly independent} in the multi-teacher distillation framework, which means that we cannot guarantee $\mathbf{p}^{\mathrm{t}}(\mathbf{x}_i)\!=\!\sum_{k=1}^K \lambda_\gamma(k,i)\widetilde{\mathbf{h}}_i^{(k)}\!=\!\mathbf{p}^*(\mathbf{x}_i)$ for node $v_i\in\mathcal{V}$ by just adjusting weights $\{\lambda_\gamma(k,i)\}_{k=1}^K$. In practice, the following Proposition \ref{theorem:1} indicates that an imperfect teacher $\mathbf{p}^{\mathrm{t}}(\mathbf{x})\neq\mathbf{p}^*(\mathbf{x})$ can still improve generalization by approximating the Bayesian teacher $\mathbf{p}^*(\mathbf{x})$.

\vspace{-0.5em}
\begin{theorem} \label{theorem:1}
Consider an integrated teacher  $\mathbf{p}^{\mathrm{t}}(\mathbf{x})$ and a Bayesian teacher $\mathbf{p}^*(\mathbf{x})$. For any GNN encoder $f_\theta(\cdot)$ and prediction head $g_\omega(\cdot)$, the difference between the distillation objective $\widetilde{R}(\theta,\omega)$ and Bayesian objective $R(\theta,\omega)$ is bounded by the Mean Square Error (MSE) of their probabilities,
\vspace{-0.3em}
\begin{equation}
\mathbb{E}\Big[\Big(\widetilde{R}(\theta,\omega)-R(\theta,\omega)\Big)^{2}\Big] \leq \frac{1}{N} \mathbb{V}\Big[\mathbf{p}^{\mathrm{t}}(\mathbf{x})^{T} \mathbf{l}\big(g_\omega(f_\theta(\mathbf{x}))\big)\Big]+\mathcal{O}\Big(\mathbb{E}\big[\|\mathbf{p}^{\mathrm{t}}(x)-\mathbf{p}^{*}(x)\|_2\big]\Big)^2
\label{equ:6}
\end{equation}
\end{theorem}
\vspace{-0.5em}
where the derivation of Eq.~(\ref{equ:6}) is available in \textbf{Appendix C}. On the right-hand side of Eq.~(\ref{equ:6}), the second term dominates when $N$ is large, which suggests that the effectiveness of knowledge distillation is governed by how close the teacher probability $\mathbf{p}^{\mathrm{t}}(\mathbf{x})$ are to the Bayesian probability $\mathbf{p}^*(\mathbf{x})$. The above discussion reached a theoretical guidance \ref{remark:1} for how to integrate the knowledge.
\vspace{-0.4em}
\begin{remark} \label{remark:1}
The instance-level knowledge weights should be set (or learned) in such a way that the integrated teacher probability $\mathbf{p}^{\mathrm{t}}(\mathbf{x})$ is as close as possible to the true Bayesian probability $\mathbf{p}^*(\mathbf{x})$.
\end{remark}

\textbf{\emph{(P2)}} In practice, precisely estimating the squared error to $\mathbf{p}^*(\mathbf{x})$ is not feasible (since $\mathbf{p}^*(\mathbf{x})$ is unknown, especially for those unlabeled data), but one can estimate the quality of the teacher probability $\mathbf{p}^{\mathrm{t}}(\mathbf{x})$ on a holdout set, e.g., by computing the log-loss or squared loss over one-hot labels \citep{menon2021statistical}. This inspired us to approximately treat $\mathbf{p}^*(\mathbf{x})\approx\mathbf{e}_y$ on the training set and optimize $\lambda_\gamma(\cdot,\cdot)$ by minimizing the cross-entropy loss $\mathcal{L}_W=\frac{1}{|\mathcal{V}_L|}\sum_{i\in\mathcal{V}_L}
\ell(\mathbf{p}^{\mathrm{t}}(\mathbf{x}),\mathbf{p}^*(\mathbf{x}))$. The learned $\lambda_\gamma(\cdot,\cdot)$ is then used to infer the proper teacher probability $\mathbf{p}^{\mathrm{t}}(\mathbf{x}_i)$ for unlabeled data $v_i\in\mathcal{V}_U$. While such estimations are often imperfect, they help to detect poor teacher probabilities, especially for those unlabeled data. The mean squared errors over one-hot labels on the training and testing sets in Fig.~\ref{fig:7} also demonstrate the effectiveness of such estimations when $\mathbf{p}^*(\mathbf{x})$ is unknown in practice.

\vspace{0.3em}
\textbf{\emph{(P3)}} Furthermore, we derive the following Theorem \ref{definition:1}, a theoretical justification for the advantages of \texttt{AGSSL} under the multi-task learning setting, which theoretically proves that the \textbf{optimal} integrated teacher $\mathbf{p}^{\mathrm{t}}(x)$ can monotonically approximate $\mathbf{p}^{*}(x)$ as the number of teachers $K$ increases.
\begin{definition} \label{definition:1}
\vspace{-0.5em}
Define $\Delta(K) \!=\! \min\big\|\mathbf{p}^{\mathrm{t}}(\mathbf{x}_i)-\mathbf{p}^{*}(\mathbf{x}_i)\big\|_2 \!=\! \min\big\|\sum_{k=1}^K \lambda_\gamma(k,i)\widetilde{\mathbf{h}_i}^{(k)}\!-\!\mathbf{p}^{*}(\mathbf{x}_i)\big\|_2$ with $K (K \geq 1)$ given teachers, then we have (1) $\Delta(K+1) \leq \Delta(K)$, and (2) $\lim _{K \rightarrow \infty} \Delta(K)=0$.
\end{definition}
\vspace{-0.3em}
where the above derivation is avaliable in \textbf{Appendix D}. The theorem \ref{definition:1} indicates \texttt{AGSSL} is endowed with the theoretical potential to benefit from more teachers, i.e., it has advantages in handling the task-level compatibility, which is also supported by the experimental results in Sec.~\ref{sec:4.3} and Fig.~\ref{fig:8}.

\subsection{Parameterized Knowledge Integration}

\begin{wrapfigure}{r}{0.4\textwidth}
\vspace{-4em}
\begin{center}
\includegraphics[width=0.4\textwidth]{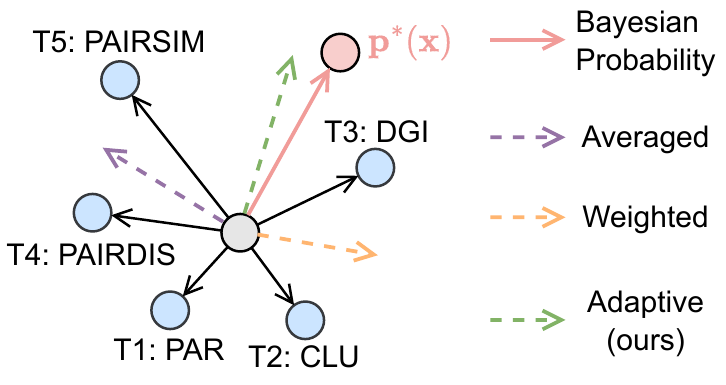}   
\end{center}
\vspace{-1em}
\caption{Illustration of the (2D) teacher probability directions for three schemes.}
\vspace{-1em}
\label{fig:4}
\end{wrapfigure}

The most common schemes for integrating different levels of knowledge from multiple teachers are averaged assignment (\texttt{Average}) or weighted integration based on labeled data (\texttt{Weighted}). However, averaged assignment fails to differentiate important teachers from irrelevant ones, and the weighted integration may mislead the student in the presence of low-quality teachers. An intuitive illustration of this issue is provided in Fig.~\ref{fig:4}, where the integrated teacher probability $\mathbf{p}^{\mathrm{t}}(\mathbf{x})$ obtained by the averaged and weighted schemes not only does not come close but even deviates from the true Bayesian probability $\mathbf{p}^*(\mathbf{x})$. Compared to the above two heuristic schemes, this paper aims to adaptively learn the knowledge integration weights by a parameterized weighting function $\lambda_\gamma(\cdot,\cdot)$.

A natural solution to achieve localized knowledge integration is to introduce a weighting function $\lambda_\gamma\left(\cdot \mid \gamma_{i}\right)$ parameterized by $\gamma_{i} \in \mathbb{R}^F$. However, directly fitting each $\lambda_\gamma\left(\cdot \mid \gamma_{i}\right)$ ($1\leq i \leq N$) locally involves solving $NF$ parameters, which increases the over-fitting risk given the limited labels in the graph. Thus, we consider the amortization inference \citep{kingma2013auto} which avoids the optimization of parameter $\gamma_{i}$ for each node locally and instead fits a shared neural network. In this section, we introduce two schemes, \texttt{AGSSL-LF} and \texttt{AGSSL-TS}, to parameterize the weighting function $\lambda_\gamma(\cdot,\cdot)$, resulting in two specific instantiations of our proposed \texttt{AGSSL} framework.

\textbf{\texttt{AGSSL-LF}}. To explicitly capture the localized importance of different teachers, we introduce a set of latent variables $\{\boldsymbol{\mu}_k\}_{k=1}^K$ and associate each teacher with a latent factor $\boldsymbol{\mu}_k\in\mathbb{R}^C$ to represent it. This strategy is inspired by latent factor models commonly applied in the recommender system \citep{koren2008factorization}, where each user or item corresponds to one latent factor used to summarize
their implicit features. The importance weight of the $k$-th teacher to node $v_i$ can be calculated as follows
\begin{equation}
    \lambda_\gamma(k, i) = \frac{\exp \left(\zeta_{k, i}\right)}{\sum_{k^{\prime}=1}^{K} \exp \left(\zeta_{k^{\prime}, i}\right)}, \quad \text{where} \textbf{  } \textbf{  } \textbf{  } \zeta_{k, i} = \boldsymbol{\nu} ^T\big(\boldsymbol{\mu}_k \odot g_\omega(f_\theta(\mathcal{G}, i))\big)
\end{equation}
where $\boldsymbol{\nu} \in \mathbb{R}^C$ is a a global parameter vector to be learned, which determines whether or not the value of each dimension in $\big(\boldsymbol{\mu}_k \odot g_\omega(f_\theta(\mathcal{G}, i))\big)$ has a positive effect on the importance score. Larger $\lambda_\gamma(k, i)$ denotes that the knowledge extracted by $k$-th teacher is more important to node $v_i$.

\textbf{\texttt{AGSSL-TS}}. Unlike \texttt{AGSSL-LF}, which calculates importance weights based solely on the node embeddings of different teachers, \texttt{AGSSL-TS} takes into account the matching degree of each teacher-student pair to distill the most matched teacher knowledge into the student model. We separately project the node logits of the student $g_\omega(f_\theta(\mathcal{G}, i))\in\mathbb{R}^C$ and each teacher $g_{\omega^*}(f_{\theta_k^*}(\mathcal{G}, i))\in\mathbb{R}^C$ into two subspaces via a linear transformation with the parameter matrix $\mathbf{W} \in \mathbb{R}^{C \times C}$. Then, the importance weight of $k$-th teacher (e.g., pretext task) to node $v_i$ can be calculated as follows
\begin{equation}
\lambda_\gamma(k, i) = \frac{\exp \left(\zeta_{k, i}\right)}{\sum_{k^{\prime}=1}^{K} \exp \left(\zeta_{k^{\prime}, i}\right)}, \quad \text{where} \textbf{  } \textbf{  } \textbf{  } \zeta_{k, i} = \Big(\mathbf{W}g_\omega(f_\theta(\mathcal{G}, i))\Big)^{T} \Big(\mathbf{W}g_{\omega_{k}^*}(f_{\theta_k^*}(\mathcal{G}, i))\Big)
\end{equation}

The pseudo-code of the proposed \texttt{AGSSL} framework is summarized in Algorithm~\ref{algo:1} in \textbf{Appendix E}.

\vspace{-0.5em}
\section{Experimental Evaluation}
In this section, we evaluate \texttt{AGSSL} on eight real-world datasets by answering the following five questions. \textbf{Q1}: Can \texttt{AGSSL} achieve better performance compared to training with individual tasks? \textbf{Q2:} How does \texttt{AGSSL} compare to other leading graph SSL baselines? \textbf{Q3:} Can \texttt{AGSSL} learn localized and customized SSL task strategies? \textbf{Q4:} Can \texttt{AGSSL} learn high-quality teacher probabilities $\mathbf{p}^{\mathrm{t}}(\mathbf{x})$? \textbf{Q5:} How do the performance of \texttt{AGSSL-LF} and \texttt{AGSSL-TS} compare with other heuristics knowledge integration approaches? Can \texttt{AGSSL} consistently benefit from multiple teachers?

\textbf{Dataset.} 
The effectiveness of the \texttt{AGSSL} framework is evaluated on eight real-world graph datasets, including Cora \citep{sen2008collective}, Citeseer \citep{giles1998citeseer}, Pubmed \citep{mccallum2000automating}, Coauthor-CS, Coauthor-Physics, Amazon-Photo, Amazon-Computers \citep{shchur2018pitfalls}, and ogbn-arxiv \citep{hu2020open}. A statistical overview of these eight datasets is placed in \textbf{Appendix F}. Each set of experiments is run five times with different random seeds, and the average accuracy and standard deviation are reported as performance metrics. Due to space limitations, we defer the implementation details and the best hyperparameter settings for each dataset to \textbf{Appendix G}.

\textbf{Baseline.} 
To evaluate the capability of \texttt{AGSSL} in automated pretext tasks combinatorial search, we followed \cite{jin2021automated} to consider five classical tasks (1) PAR \citep{you2020does}, which predicts pseudo-labels from graph partitioning; (2) CLU \citep{you2020does}, which predicts pseudo-labels from $K$-means clustering on node features; (3) DGI \citep{velickovic2019deep}, which maximizes the mutual information between graph and node representations; (4) PAIRDIS \citep{jin2020self}, which predicts the shortest path length between nodes; and (5) PAIRSIM \citep{jin2020self}, which predicts the feature similarity between nodes. The detailed methodologies for the above five pretext tasks and the reasons why we selected them can be found in \textbf{Appendix H}. In addition to comparing with the automated Graph SSL - AutoSSL \citep{jin2021automated}, we also compare \texttt{AGSSL} with some of the state-of-the-art self-supervised baselines in Table.~\ref{tab:2}, including GMI \citep{peng2020graph}, MVGRL \citep{hassani2020contrastive}, GRACE \citep{zhu2020deep}, GCA \citep{zhu2020graph}, CG3 \citep{wan2020contrastive}, and BGRL \citep{thakoor2021bootstrapped}. Note that (1) there is no conflict at all between Graph SSL automation and designing more powerful pretext tasks; and (2) as a general framework, \texttt{AGSSL} is applicable to other more complex self-supervised tasks, which will be left for future work.

\begin{table*}[!htbp]
\begin{center}
\vspace{-1em}
\caption{Performance comparison of single- and multi-task learning, where \textbf{bold} and \underline{underline} denote the best metrics in multi- and single-task learning. Besides, we mark those metrics in multi-task learning that are poorer to vanilla GCNs and (the best) single-task learning as \textcolor{red}{red} and \textcolor{blue}{blue}.}
\label{tab:1}
\resizebox{\textwidth}{!}{
\begin{tabular}{clclllll|llll}

\toprule
\multirow{2}{*}{\textbf{Dataset}} &
  \multirow{2}{*}{\textbf{Setting}} &
  \multirow{2}{*}{\texttt{GCNs}} &
  \multicolumn{5}{c}{\textbf{Single Self-Supervised Task Learning}} &
  \multicolumn{4}{|c}{\textbf{Multi Self-Supervised Task Learning}} \\ \cmidrule(l){4-12} 
                              &     &            & \texttt{PAR}        & \texttt{CLU}        & \texttt{DGI}        & \texttt{PAIRDIS}    & \texttt{PAIRSIM}    & \texttt{Vanilla}    & \texttt{AutoSSL}       & \cellcolor{gray!20}\texttt{AGSSL-LF}         & \cellcolor{gray!20}\texttt{AGSSL-TS}         \\ \midrule
\multirow{2}{*}{\texttt{Cora}}         & JT & $81.72_{\pm0.52}$ & $\textit{\underline{83.52}}_{\pm0.39}$ & $82.34_{\pm0.46}$ & $83.28_{\pm0.33}$ & $82.92_{\pm0.41}$ & $83.16_{\pm0.38}$ & \textcolor{red}{$81.50_{\pm0.40}$} & $83.78_{\pm0.45}$ & \cellcolor{gray!20}$84.68_{\pm0.39}$ & \cellcolor{gray!20}$\textbf{85.32}_{\pm0.32}$ \\
                              & P\&F & $81.72_{\pm0.52}$ & $82.38_{\pm0.31}$ & $81.42_{\pm0.35}$ & $82.10_{\pm0.44}$ & $81.92_{\pm0.42}$ & $\textit{\underline{82.44}}_{\pm0.36}$ & \textcolor{red}{$80.74_{\pm0.38}$} & $82.96_{\pm0.43}$ & \cellcolor{gray!20}$84.22_{\pm0.28}$ & \cellcolor{gray!20}$\textbf{84.38}_{\pm0.27}$ \\ \midrule
\multirow{2}{*}{\texttt{Citeseer}}     & JT & $71.48_{\pm0.46}$ & $72.72_{\pm0.36}$ & $72.14_{\pm0.50}$ & $73.08_{\pm0.45}$ & $\textit{\underline{73.16}}_{\pm0.42}$ & $72.90_{\pm0.45}$ & \textcolor{blue}{$72.30_{\pm0.50}$} & $73.30_{\pm0.37}$ & \cellcolor{gray!20}$\textbf{74.34}_{\pm0.31}$ & \cellcolor{gray!20}$74.20_{\pm0.42}$ \\
                              & P\&F & $71.48_{\pm0.46}$ & $72.36_{\pm0.58}$ & $71.84_{\pm0.49}$ & $\textit{\underline{72.52}}_{\pm0.37}$ & $72.22_{\pm0.53}$ & $71.98_{\pm0.62}$ & \textcolor{blue}{$71.64_{\pm0.49}$} & $72.76_{\pm0.44}$ & \cellcolor{gray!20}$73.58_{\pm0.56}$ & \cellcolor{gray!20}$\textbf{73.70}_{\pm0.76}$ \\ \midrule
\multirow{2}{*}{\texttt{Pubmed}}       & JT & $79.26_{\pm0.40}$ & $\textit{\underline{82.16}}_{\pm0.54}$ & $80.92_{\pm0.36}$ & $81.50_{\pm0.43}$ & $81.22_{\pm0.55}$ & $80.50_{\pm0.54}$ & \textcolor{blue}{$80.86_{\pm0.50}$} & $82.72_{\pm0.35}$ & \cellcolor{gray!20}$82.66_{\pm0.32}$ & \cellcolor{gray!20}$\textbf{82.82}_{\pm0.29}$ \\
                              & P\&F & $79.26_{\pm0.40}$ & $79.56_{\pm0.39}$ & $79.12_{\pm0.47}$ & $\textit{\underline{79.90}}_{\pm0.52}$ & $79.64_{\pm0.48}$ & $79.34_{\pm0.60}$ & \textcolor{red}{$78.90_{\pm0.54}$} & $80.14_{\pm0.41}$ & \cellcolor{gray!20}$\textbf{80.62}_{\pm0.25}$ & \cellcolor{gray!20}$80.54_{\pm0.42}$ \\ \midrule
\multirow{2}{*}{\texttt{CS}}  & JT & $91.04_{\pm0.45}$ & $92.30_{\pm0.67}$ & $92.94_{\pm0.70}$ & $92.66_{\pm0.69}$ & $92.48_{\pm0.55}$ & $\textit{\underline{93.12}}_{\pm0.64}$ & \textcolor{blue}{$92.16_{\pm0.60}$} & $93.54_{\pm0.46}$ & \cellcolor{gray!20}$\textbf{93.86}_{\pm0.36}$ & \cellcolor{gray!20}$93.46_{\pm0.25}$ \\
                              & P\&F & $91.04_{\pm0.45}$ & $91.28_{\pm0.55}$ & $91.36_{\pm0.63}$ & $\textit{\underline{91.80}}_{\pm0.73}$ & $91.44_{\pm0.49}$ & $91.62_{\pm0.47}$ & \textcolor{blue}{$91.42_{\pm0.57}$} & $\textbf{92.48}_{\pm0.45}$ & \cellcolor{gray!20}$92.36_{\pm0.45}$ & \cellcolor{gray!20}$91.94_{\pm0.33}$ \\ \midrule
\multirow{2}{*}{\texttt{Physics}} & JT & $93.06_{\pm0.55}$ & $94.08_{\pm0.56}$ & $94.12_{\pm0.49}$ & $\textit{\underline{94.74}}_{\pm0.46}$ & $94.62_{\pm0.63}$ & $94.40_{\pm0.48}$ & \textcolor{blue}{$93.94_{\pm0.47}$} & $95.10_{\pm0.42}$ & \cellcolor{gray!20}$\textbf{95.74}_{\pm0.38}$ & \cellcolor{gray!20}$95.54_{\pm0.35}$ \\
                              & P\&F & $93.06_{\pm0.55}$ & $93.18_{\pm0.71}$ & $93.50_{\pm0.53}$ & $93.92_{\pm0.60}$ & $\textit{\underline{94.04}}_{\pm0.56}$ & $93.34_{\pm0.73}$ & \textcolor{blue}{$93.40_{\pm0.50}$} & \textcolor{blue}{$93.88_{\pm0.45}$} & \cellcolor{gray!20}$94.80_{\pm0.29}$ & \cellcolor{gray!20}$\textbf{94.96}_{\pm0.43}$ \\ \midrule
\multirow{2}{*}{\texttt{Photo}} & JT & $91.90_{\pm0.46}$ & $92.54_{\pm0.60}$ & $\textit{\underline{93.04}}_{\pm0.55}$ & $92.46_{\pm0.70}$ & $92.32_{\pm0.55}$ & $92.82_{\pm0.78}$ & \textcolor{red}{$91.52_{\pm0.61}$} & \textcolor{blue}{$92.94_{\pm0.40}$} & \cellcolor{gray!20}$93.98_{\pm0.29}$ & \cellcolor{gray!20}$\textbf{94.22}_{\pm0.31}$ \\
                              & P\&F & $91.90_{\pm0.46}$ & $92.24_{\pm0.49}$ & $\textit{\underline{92.58}}_{\pm0.66}$ & $92.02_{\pm0.59}$ & $92.10_{\pm0.52}$ & $92.42_{\pm0.44}$ & \textcolor{red}{$90.84_{\pm0.51}$} & \textcolor{blue}{$92.36_{\pm0.45}$} & \cellcolor{gray!20}$93.32_{\pm0.37}$ & \cellcolor{gray!20}$\textbf{93.52}_{\pm0.41}$ \\ \midrule
\multirow{2}{*}{\texttt{Computers}}   & JT & $86.36_{\pm0.65}$ & $87.48_{\pm0.65}$ & $87.96_{\pm0.72}$ & $88.08_{\pm0.64}$ & $87.62_{\pm0.52}$ & $\textit{\underline{88.40}}_{\pm0.72}$ & \textcolor{blue}{$86.58_{\pm0.50}$} & $88.72_{\pm0.44}$ & \cellcolor{gray!20}$89.56_{\pm0.34}$ & \cellcolor{gray!20}$\textbf{89.72}_{\pm0.28}$ \\
                              & P\&F & $86.36_{\pm0.65}$ & $86.72_{\pm0.78}$ & $\textit{\underline{87.74}}_{\pm0.80}$ & $87.36_{\pm0.73}$ & $86.52_{\pm0.65}$ & $87.20_{\pm0.69}$ & \textcolor{red}{$85.90_{\pm0.57}$} & $88.00_{\pm0.49}$ & \cellcolor{gray!20}$\textbf{88.68}_{\pm0.42}$ & \cellcolor{gray!20}$88.42_{\pm0.33}$ \\ \midrule
\multirow{2}{*}{\texttt{ogbn-arxiv}}   & JT & $71.16_{\pm0.32}$ & $71.84_{\pm0.28}$ & $71.72_{\pm0.40}$ & $72.04_{\pm0.25}$ & $\textit{\underline{72.18}}_{\pm0.30}$ & $71.90_{\pm0.33}$ & \textcolor{red}{$70.94_{\pm0.33}$} & $72.26_{\pm0.25}$ & \cellcolor{gray!20}$72.66_{\pm0.26}$ & \cellcolor{gray!20}$\textbf{72.72}_{\pm0.22}$ \\
                              & P\&F & $71.16_{\pm0.32}$ & $71.78_{\pm0.37}$ & $71.54_{\pm0.36}$ & $\textit{\underline{71.96}}_{\pm0.28}$ & $71.90_{\pm0.33}$ & $71.62_{\pm0.29}$ & \textcolor{red}{$70.56_{\pm0.31}$} & $72.08_{\pm0.24}$ & \cellcolor{gray!20}$72.52_{\pm0.31}$ & \cellcolor{gray!20}$\textbf{72.60}_{\pm0.25}$ \\ \midrule
\multirow{2}{*}{Avg. Rank $\downarrow$} & JT & 9.63       & 6.50       & 6.63       & 5.25       & 6.00       & 5.75       & 9.13       & 2.88       & \cellcolor{gray!20}1.75       & \cellcolor{gray!20}1.50       \\
                              & P\&F & 9.13       & 6.63       & 6.88       & 4.88       & 5.88       & 6.00       & 9.13       & 3.25       & \cellcolor{gray!20}1.88       & \cellcolor{gray!20}1.38       \\ \bottomrule
                              
\end{tabular}} \vspace{-1.5em}
\end{center}
\end{table*}

\subsection{Performance Comparision}
\textbf{Performance Comparision with Individual Tasks (Q1).} To answer \textbf{Q1}, we report the results for individual and multiple self-supervised tasks under two training strategies, i.e., \textit{Joint Training} (\texttt{JT}) and \textit{Pre-train\&Fine-tune} (\texttt{P\&F}) in Table.~\ref{tab:1}, from which we can make the following observations: (1) The performance of individual pretext tasks depends heavily on the datasets, and there does not exist an ``optimal" task that works for all datasets. (2) Simply averaging over all tasks (\texttt{Vanilla}) may cause a serious task-level compatibility problem, whose performance may not only be inferior to training with individual tasks (marked in \textcolor{blue}{blue}), but even poorer than the vanilla implementation of GCNs (marked in \textcolor{red}{red}). (3) As an automated SSL approach, \texttt{AutoSSL} performs better than simply averaging over all tasks, but still lags far behind our proposed \texttt{AGSSL} overall on eight datasets.

\textbf{Performance Comparision with Representative SSL Baselines (Q2).} To answer \textbf{Q2}, we compare \texttt{AGSSL} with several representative graph self-supervised baselines. As can be seen from the results reported in Table.~\ref{tab:2}, by combining just a few simple and classical pretext tasks, the resulting performance is comparable to that of several state-of-the-art self-supervised baselines. For example, \texttt{AGSSL-LF} and \texttt{AGSSL-TS} perform better than other compared methods on 5 out of 8 datasets. Note that these results are provided only to demonstrate the power of \texttt{AGSSL}, and we would like to emphasize again that \texttt{AGSSL} is general enough to be combined with any existing graph SSL tasks.

\begin{table*}[!htbp]
\begin{center}
\vspace{-1em}
\caption{Performance comparison with classical self-supervised algorithms under the \textit{Joint Training} setting, where \textbf{bold} and \underline{underline} denote the best and second metrics on each dataset, respectively.}
\label{tab:2}
\resizebox{1.0\textwidth}{!}{
\begin{tabular}{lllllllll|c}

\toprule
\textbf{Method}      & \texttt{Cora}       & \texttt{Citeseer}   & \texttt{Pubmed}     & \texttt{CS} & \texttt{Physics} & \texttt{Photo} & \texttt{Computers} & \texttt{ogbn-arxiv} & Avg. Rank $\downarrow$ \\ \midrule
\texttt{GCNs} & $81.72_{\pm0.52}$ & $71.48_{\pm0.46}$ & $79.26_{\pm0.40}$ & $91.04_{\pm0.45}$ & $93.06_{\pm0.55}$ & $91.90_{\pm0.46}$ & $86.36_{\pm0.65}$ & $71.16_{\pm0.32}$ & 9.83 \\
\texttt{DGI}         & $83.28_{\pm0.33}$ & $73.08_{\pm0.45}$ & $81.50_{\pm0.43}$ & $92.66_{\pm0.69 }$ & $94.74_{\pm0.46  }$ & $92.46_{\pm0.70  }$ & $88.08_{\pm0.64}$ & $72.04_{\pm0.25}$ & 7.50         \\
\texttt{GMI}         & $82.94_{\pm0.40}$ & $73.22_{\pm0.38}$ & $81.20_{\pm0.35}$ & $92.76_{\pm0.56 }$ & \textit{OOM} & $92.74_{\pm0.56  }$ & $88.20_{\pm0.45}$ & \textit{OOM} & 7.00         \\
\texttt{MVGRL}       & $83.36_{\pm0.43}$ & $72.66_{\pm0.37}$ & $81.74_{\pm0.41}$ & $92.84_{\pm0.39 }$ & \textit{OOM} & $93.06_{\pm0.45  }$ & $88.36_{\pm0.51}$ & \textit{OOM} & 6.33         \\
\texttt{GRACE}       & $80.80_{\pm0.38}$ & $72.24_{\pm0.44}$ & $79.96_{\pm0.46}$ & $91.94_{\pm0.37 }$ & $93.64_{\pm0.47  }$ & $91.92_{\pm0.43  }$ & $87.44_{\pm0.49}$ & \textit{OOM} & 9.17        \\
\texttt{GCA}         & $84.34_{\pm0.45}$ & $73.72_{\pm0.37}$ & $81.98_{\pm0.42}$ & $93.30_{\pm0.42 }$ & $94.78_{\pm0.52  }$ & $93.30_{\pm0.36  }$ & $88.74_{\pm0.37}$ & \textit{OOM} & 4.17         \\
\texttt{CG3}         & $83.76_{\pm0.39}$ & $73.54_{\pm0.40}$ & $81.58_{\pm0.36}$ & $93.02_{\pm0.51 }$ & $94.90_{\pm0.39  }$ & $93.68_{\pm0.48  }$ & $88.42_{\pm0.42}$ & $72.40_{\pm0.24}$ & 4.83         \\
\texttt{BGRL}        & $\textit{\underline{84.82}}_{\pm0.41}$ & $73.96_{\pm0.35}$ & $82.20_{\pm0.34}$ & $\textit{\underline{93.58}}_{\pm0.29 }$ & $95.12_{\pm0.44  }$ & $93.48_{\pm0.51  }$ & $89.08_{\pm0.38}$ & $\textbf{72.80}_{\pm0.20}$ & 2.83         \\ \midrule
\cellcolor{gray!20}\texttt{AGSSL-LF}  & \cellcolor{gray!20}$84.68_{\pm0.39}$ & \cellcolor{gray!20}$\textbf{74.34}_{\pm0.31}$ & \cellcolor{gray!20}$\textit{\underline{82.66}}_{\pm0.32}$ & \cellcolor{gray!20}$\textbf{93.86}_{\pm0.36 }$ & \cellcolor{gray!20}$\textbf{95.74}_{\pm0.38  }$ & \cellcolor{gray!20}$\textit{\underline{93.98}}_{\pm0.29  }$ & \cellcolor{gray!20}$\textit{\underline{89.56}}_{\pm0.34}$ & \cellcolor{gray!20}$72.66_{\pm0.26}$ & \cellcolor{gray!20}1.83         \\
\cellcolor{gray!20}\texttt{AGSSL-TS}  & \cellcolor{gray!20}$\textbf{85.32}_{\pm0.32}$ & \cellcolor{gray!20}$\textit{\underline{74.20}}_{\pm0.42}$ & \cellcolor{gray!20}$\textbf{82.82}_{\pm0.29}$ & \cellcolor{gray!20}$93.46_{\pm0.25 }$ & \cellcolor{gray!20}$\textit{\underline{95.54}}_{\pm0.35  }$ & \cellcolor{gray!20}$\textbf{94.22}_{\pm0.31  }$ & \cellcolor{gray!20}$\textbf{89.72}_{\pm0.28}$ & \cellcolor{gray!20}$\textit{\underline{72.72}}_{\pm0.22}$ & \cellcolor{gray!20}1.50         \\ \bottomrule

\end{tabular}} \vspace{-1em}
\end{center}
\end{table*}

\subsection{Evaluation on Localized SSL Tasks and Learning Curves}
\begin{figure*}[!tbp]
	\begin{center}
		\subfigure[AGSSL-LF on Citeseer]{\includegraphics[width=0.238\linewidth]{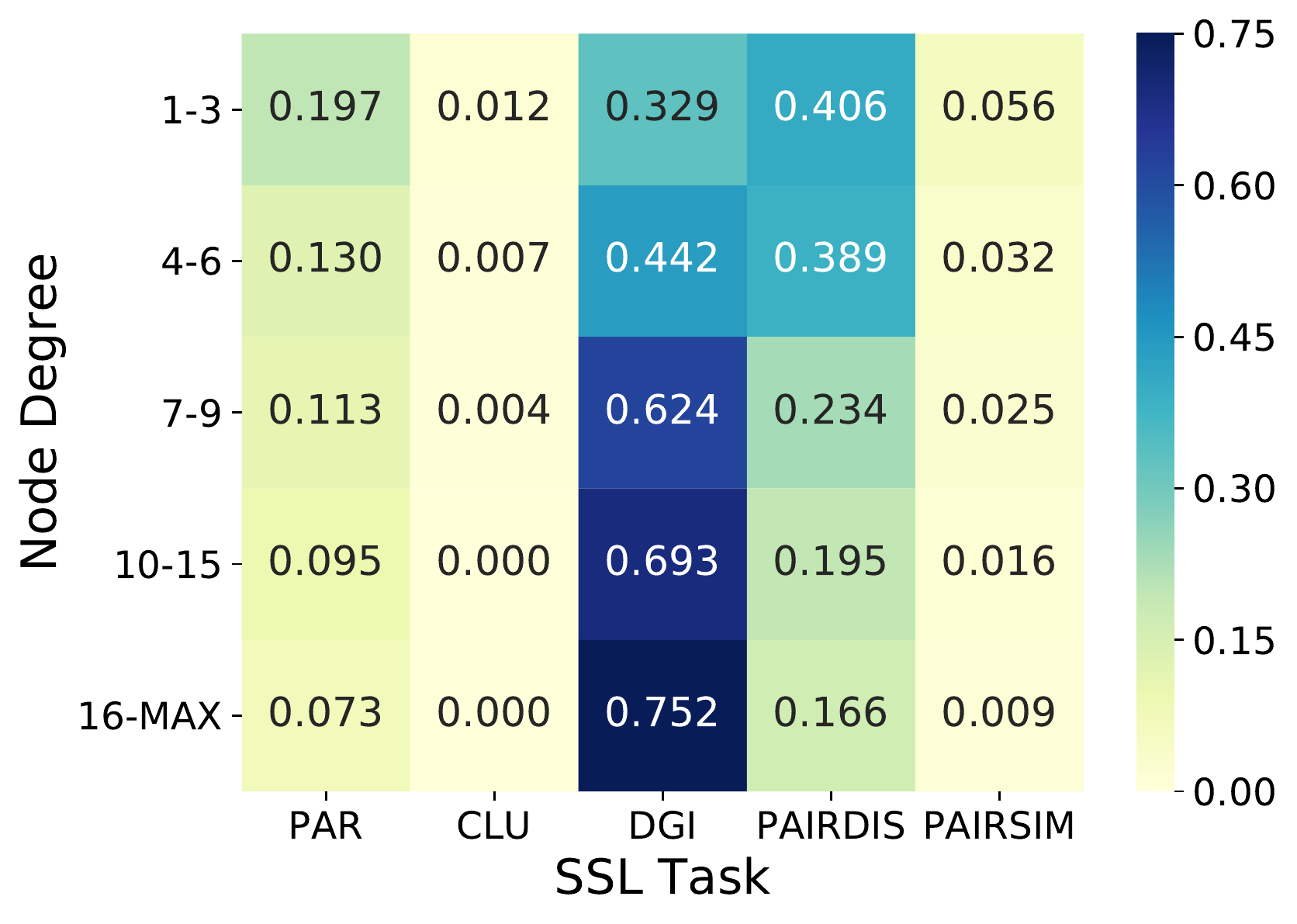} \label{fig:5a}}
		\subfigure[AGSSL-TS on Citeseer]{\includegraphics[width=0.238\linewidth]{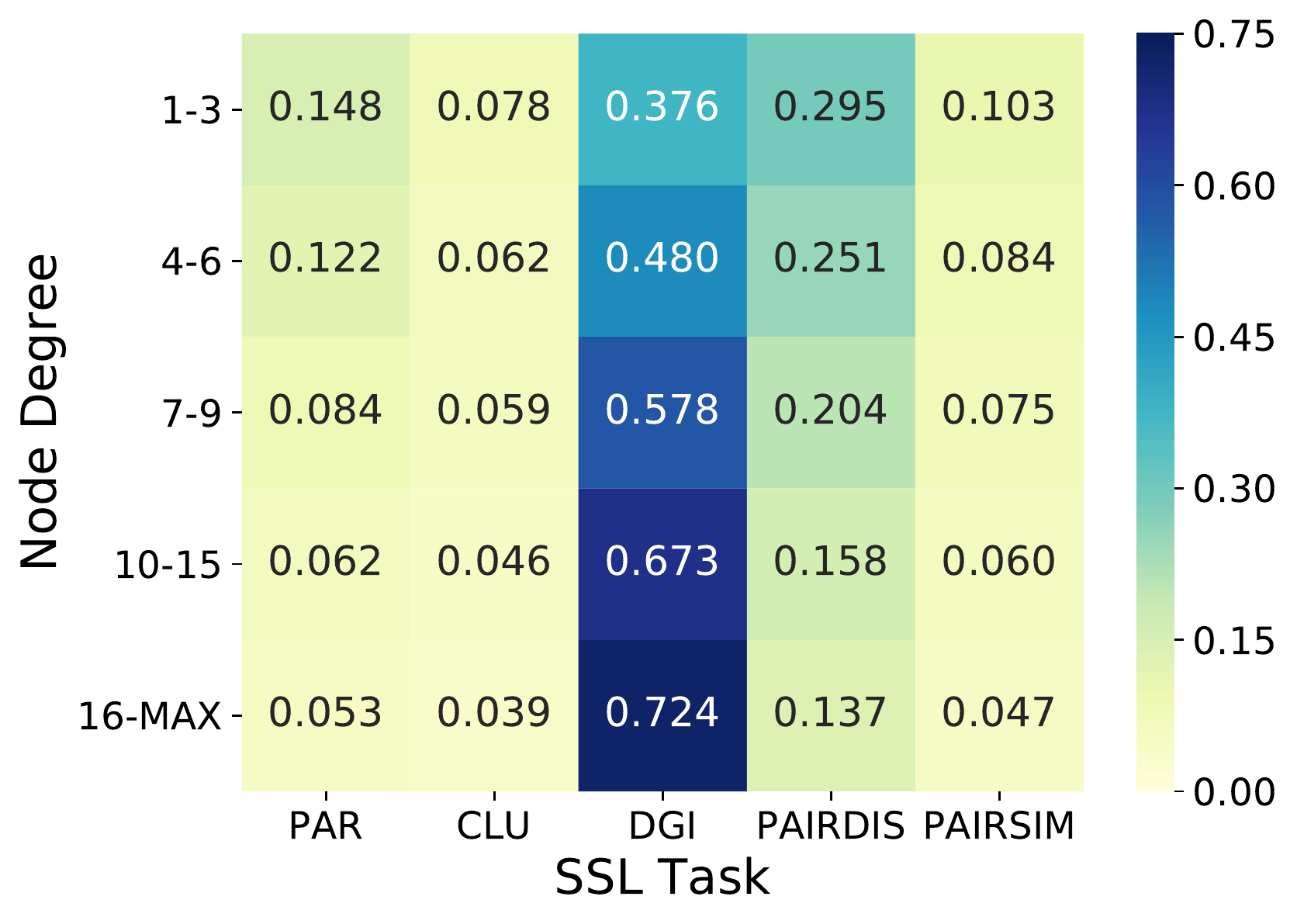} \label{fig:5b}}
		\subfigure[AGSSL-LF on CS]{\includegraphics[width=0.238\linewidth]{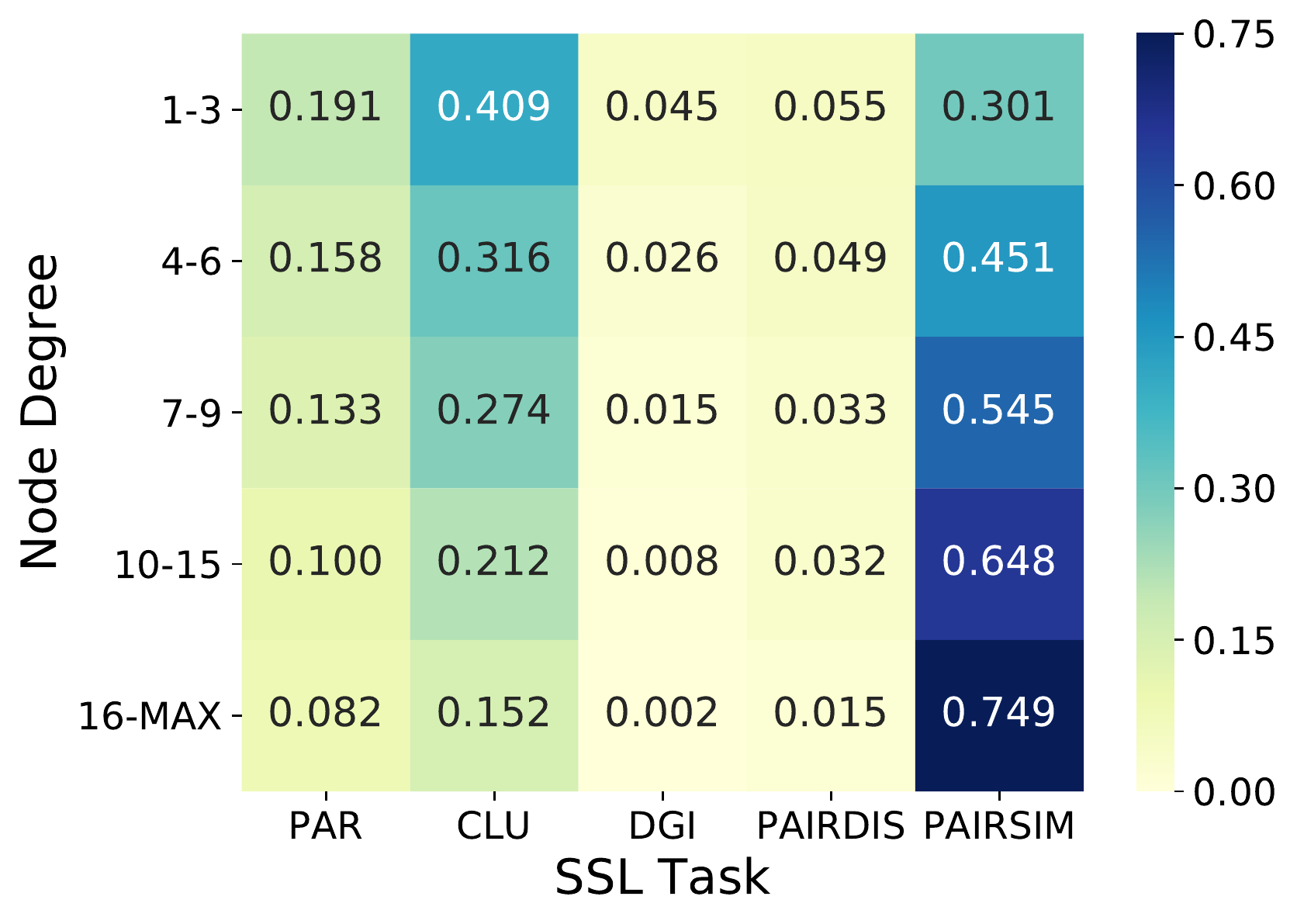} \label{fig:5c}}
		\subfigure[AGSSL-TS on CS]{\includegraphics[width=0.238\linewidth]{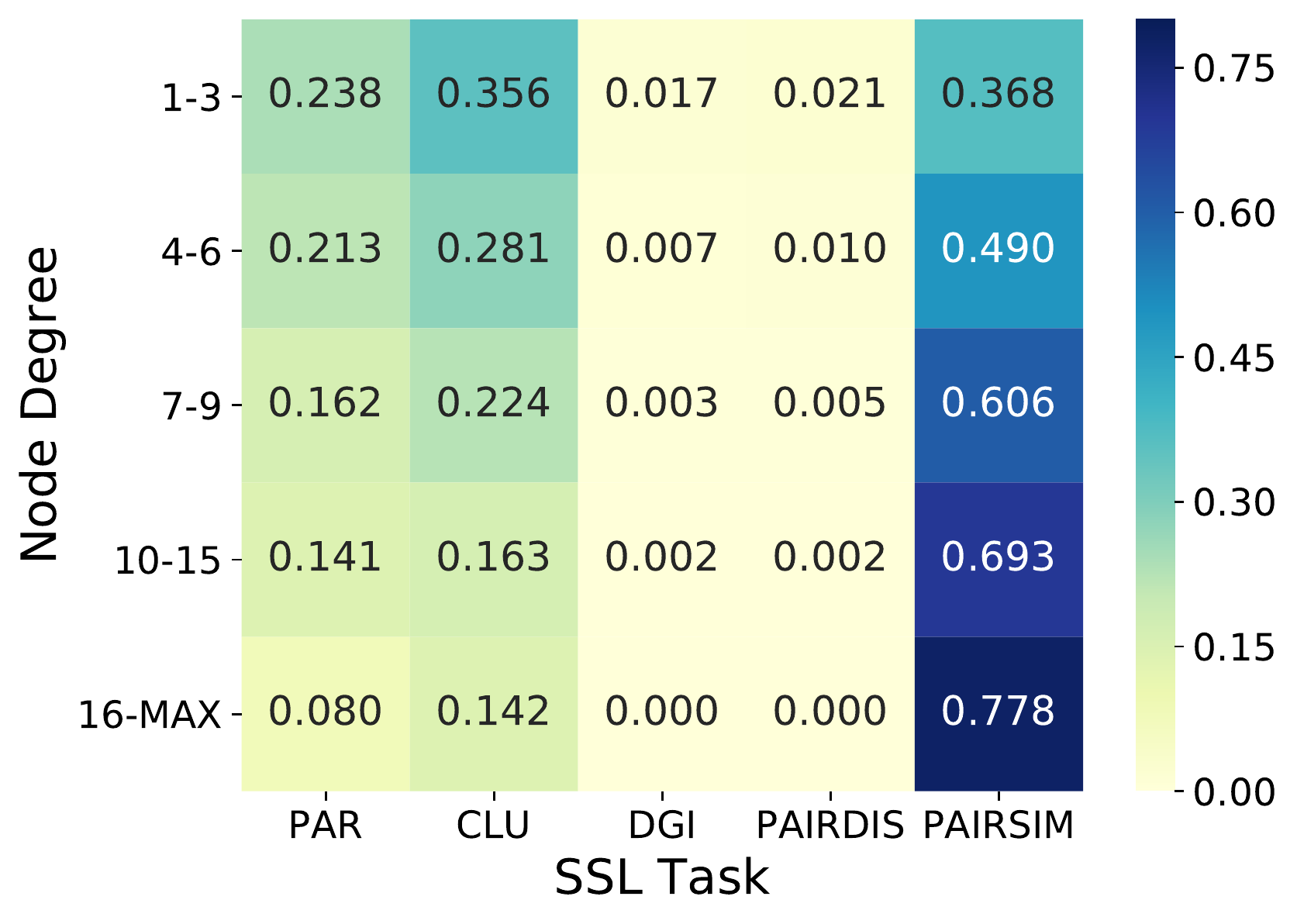} \label{fig:5d}}
	\end{center}
	\vspace{-1em}
	\caption{Illustration of average knowledge weights for nodes with different node degree ranges.}
	\vspace{-1em}
	\label{fig:5}
\end{figure*}

\vspace{-0.5em}
\textbf{Localized and Customized SSL strategies (Q3).} 
To answer \textbf{Q3}, we visualize the average knowledge weights learned by \texttt{AGSSL-LF} and \texttt{AGSSL-TS} at different node degrees on the \texttt{Citeseer} and \texttt{Coatuhor-CS} datasets. From the heatmaps shown in Fig.~\ref{fig:5}, we can make the following observations: (1) The learned weights vary a lot from dataset to dataset. For example, \texttt{Citeseer} can benefit more from pretext tasks - \texttt{DGI} and \texttt{PAIRDIS}, while pretext tasks \texttt{CLU} and \texttt{PAIRSIM} are more beneficial for \texttt{Coauthor-CS}. (2) The knowledge weights learned by \texttt{AGSSL-LF} and \texttt{AGSSL-TS} are very similar on the same dataset, suggesting that they do uncover some ``essence" rather than being completely randomized. (3) The knowledge weights vary greatly across different node degrees, and this variation is almost monotonic. For example, as the node degree increases on \texttt{Citeseer}, the dependence of nodes on \texttt{DGI} increases, while the dependence on \texttt{PAIRDIS} gradually decreases, which indicates that \texttt{AGSSL} has an advantage in learning localized SSL strategies.

\begin{figure*}[!tbp]
	\begin{center}
		\subfigure[AGSSL-LF on Citeseer]{\includegraphics[width=0.238\linewidth]{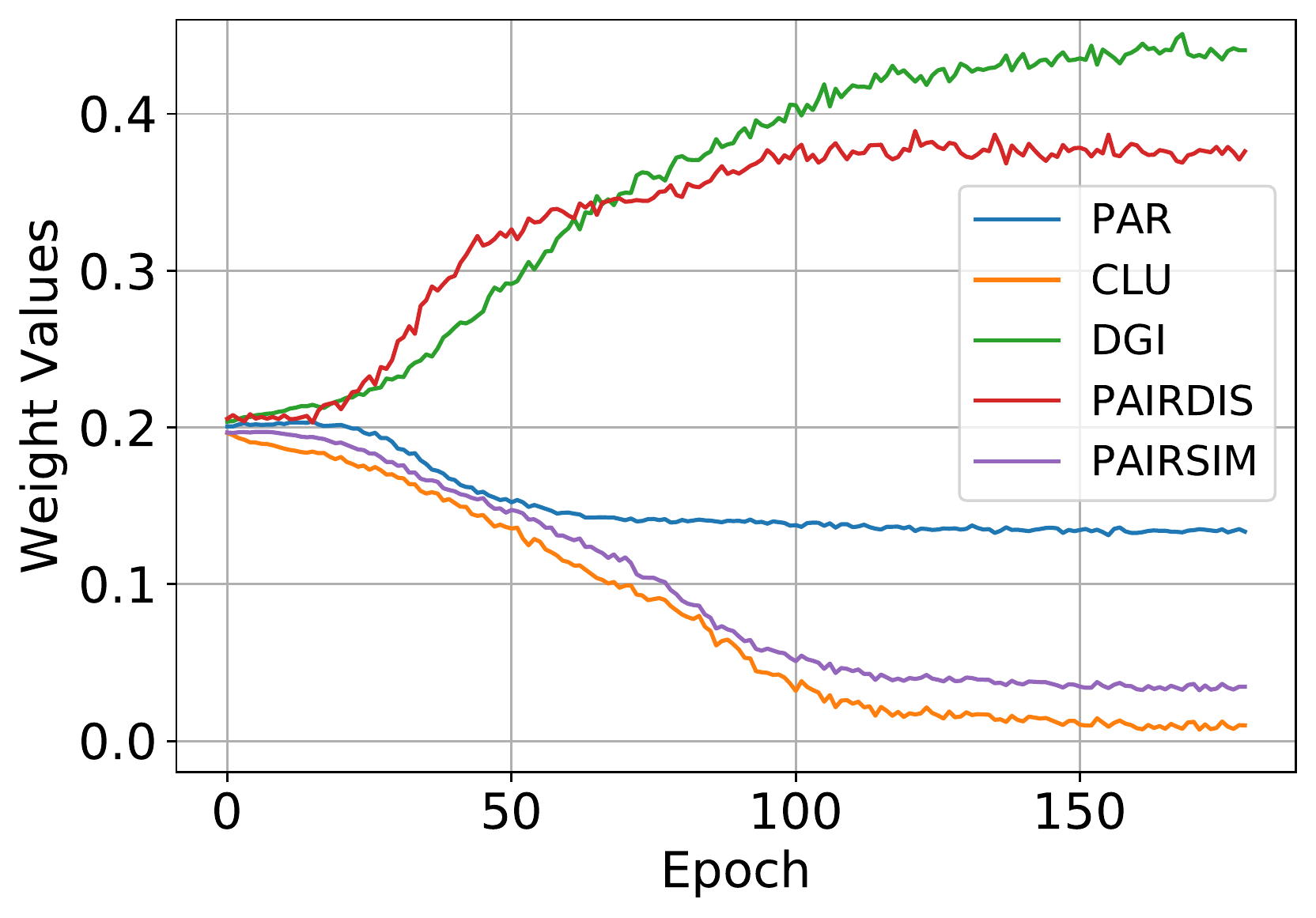} \label{fig:6a}}
		\subfigure[AGSSL-TS on Citeseer]{\includegraphics[width=0.238\linewidth]{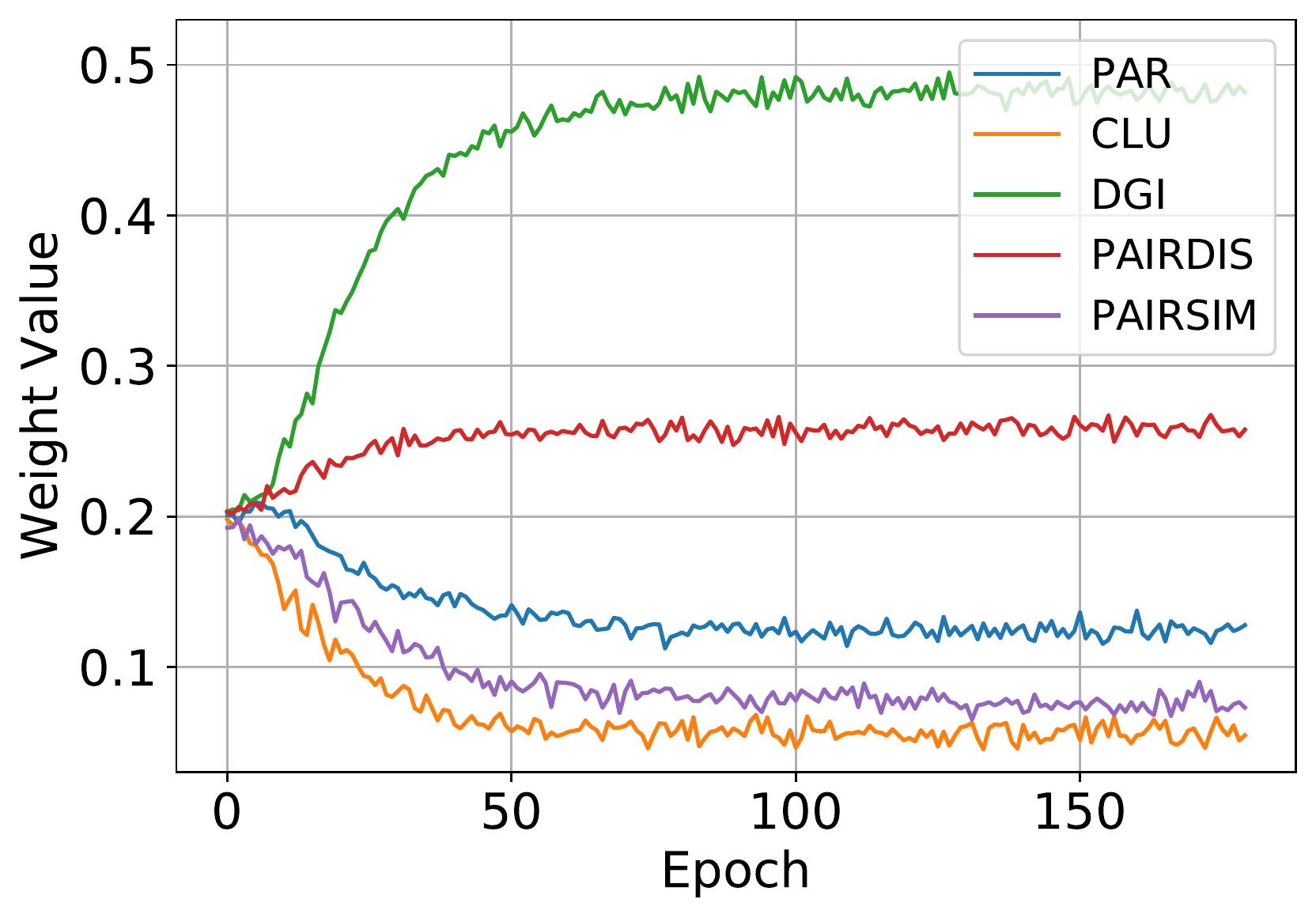} \label{fig:6b}}
		\subfigure[AGSSL-LF on CS]{\includegraphics[width=0.238\linewidth]{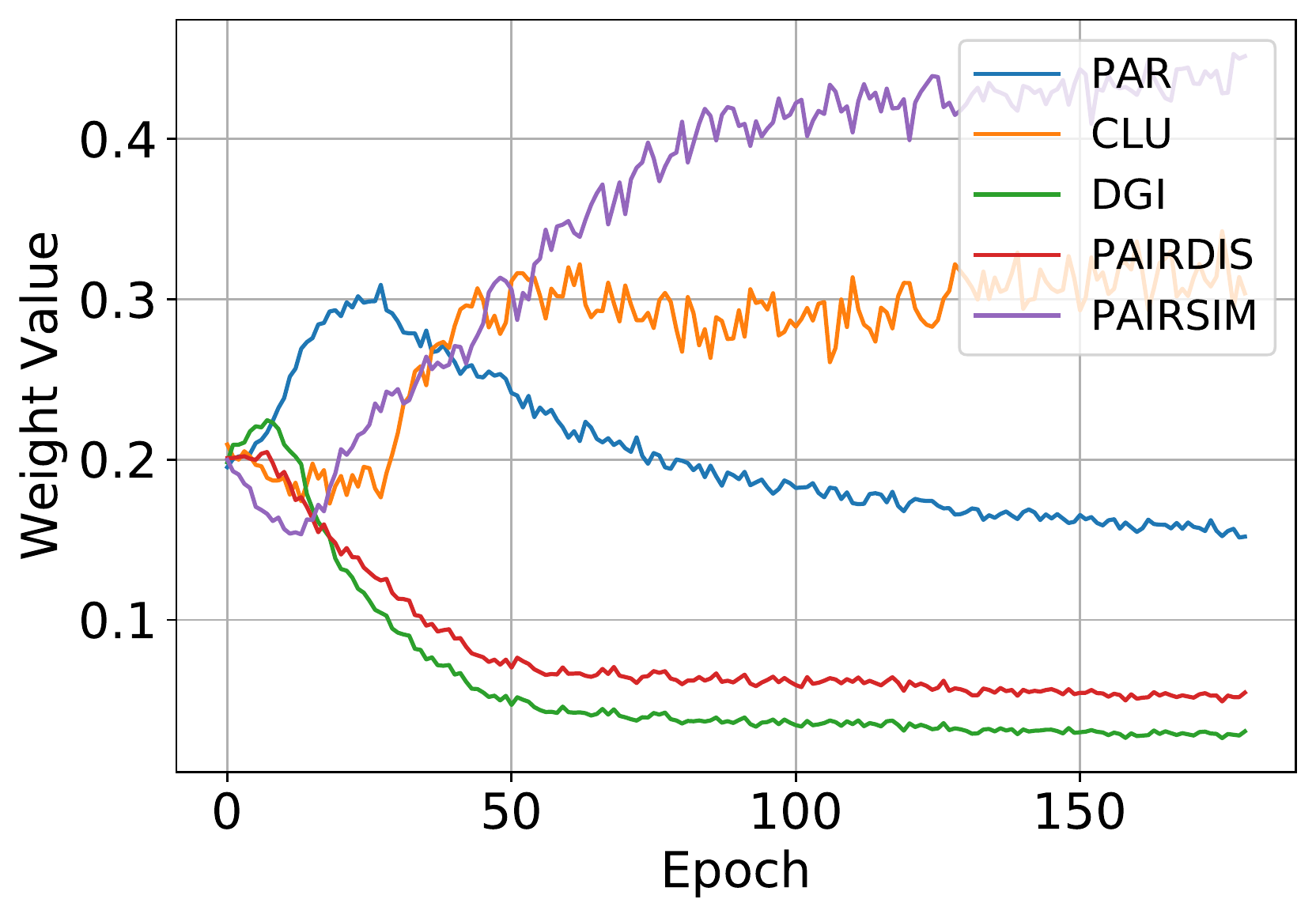} \label{fig:6c}}
		\subfigure[AGSSL-TS on CS]{\includegraphics[width=0.238\linewidth]{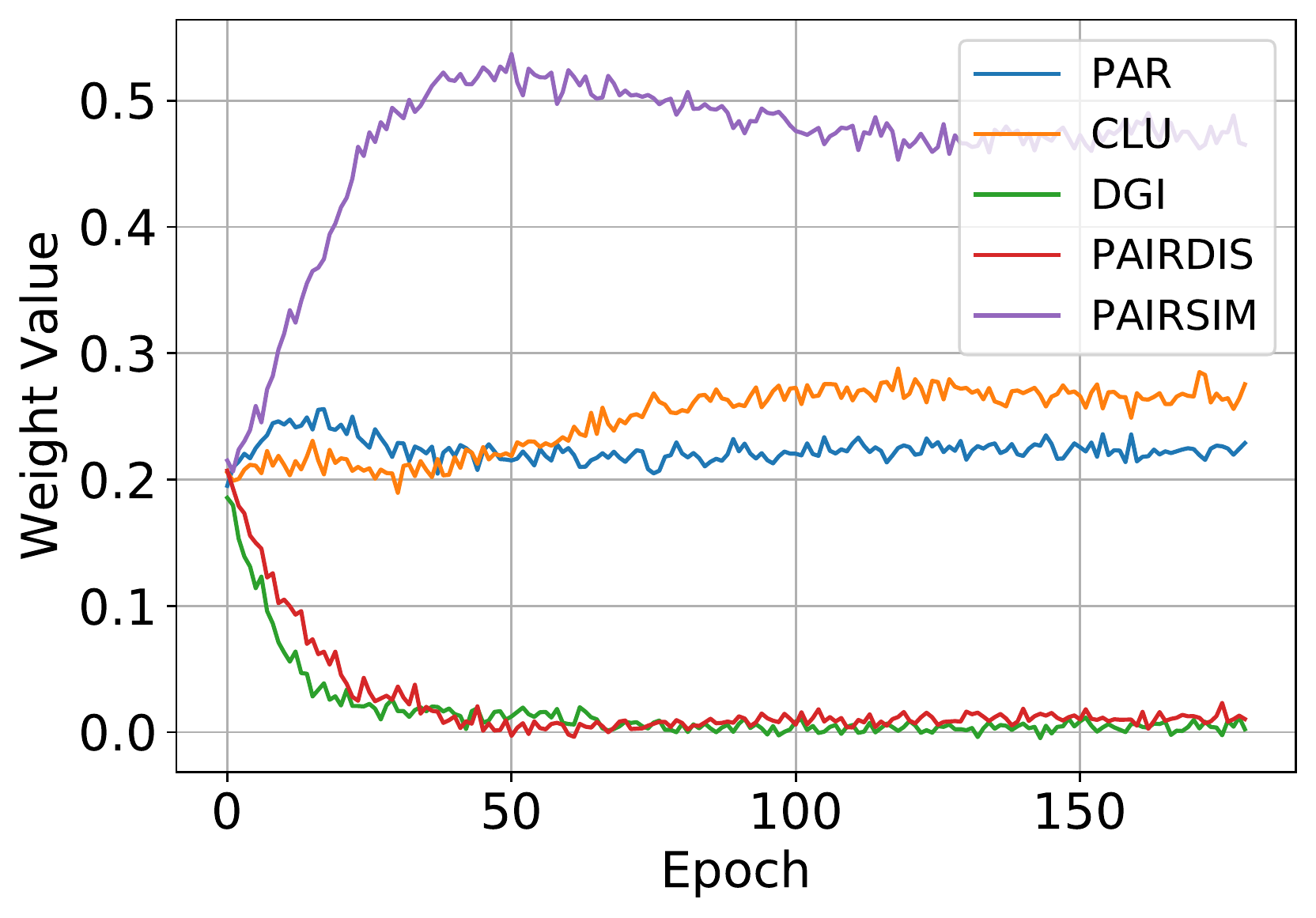} \label{fig:6d}}
	\end{center}
	\vspace{-1em}
	\caption{Evolution process of average knowledge weights for nodes with a degree range of [4, 6].}
	\vspace{-1em}
	\label{fig:6}
\end{figure*}

Furthermore, we also provide in Fig.~\ref{fig:6} the evolution process of knowledge weights for nodes with a degree range of [4, 6] on the \texttt{Citeseer} and \texttt{Coatuhor-CS} datasets. The weights of the five tasks eventually become stable and converge to steady values, corresponding to the results in Fig.~\ref{fig:5}.

\begin{figure*}[!tbp]
    \vspace{-1em}
	\begin{center}
		\subfigure[AGSSL-LF on Citeseer]{\includegraphics[width=0.238\linewidth]{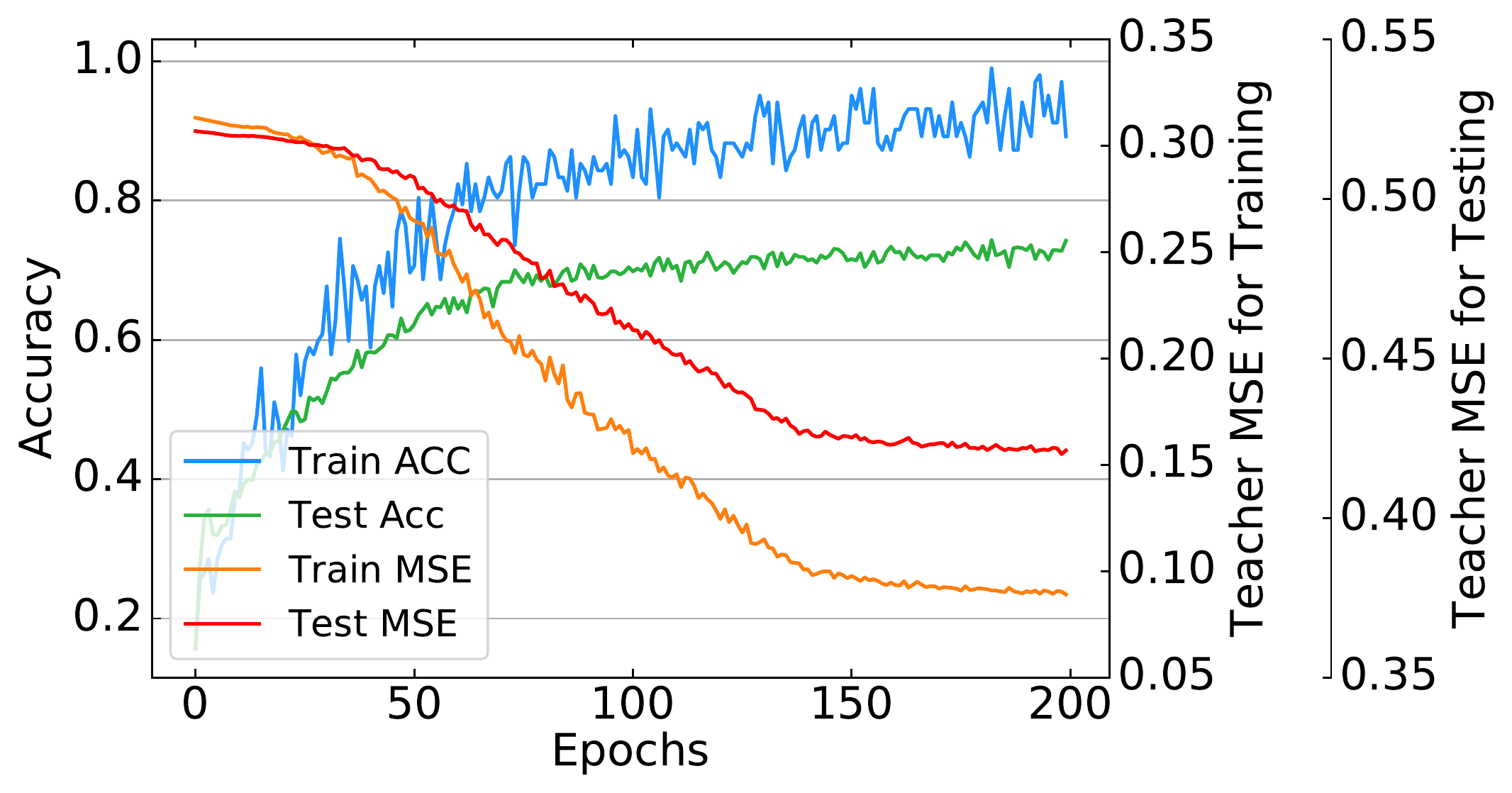} \label{fig:7a}}
		\subfigure[AGSSL-TS on Citeseer]{\includegraphics[width=0.238\linewidth]{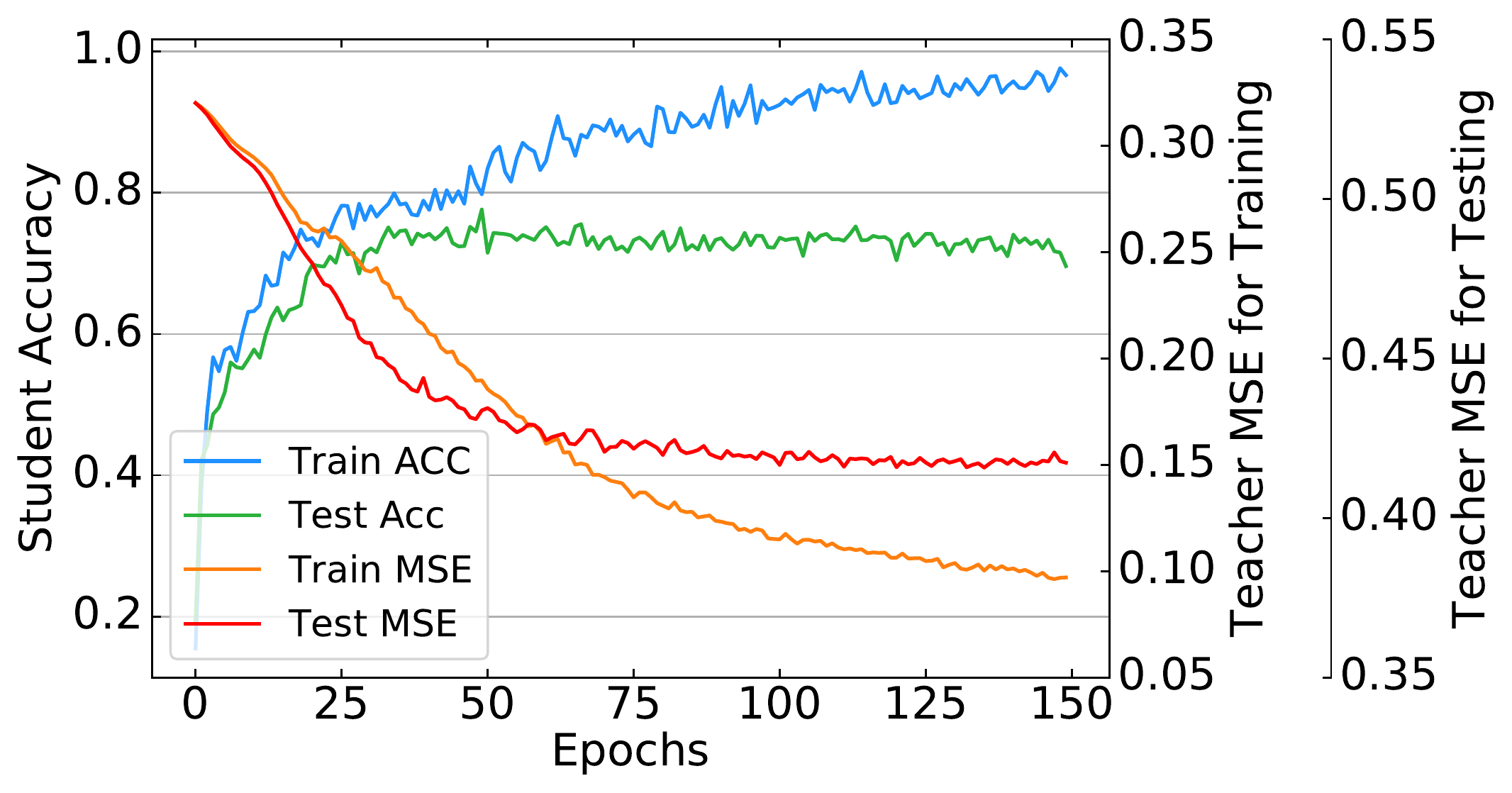} \label{fig:74b}}
		\subfigure[AGSSL-LF on CS]{\includegraphics[width=0.238\linewidth]{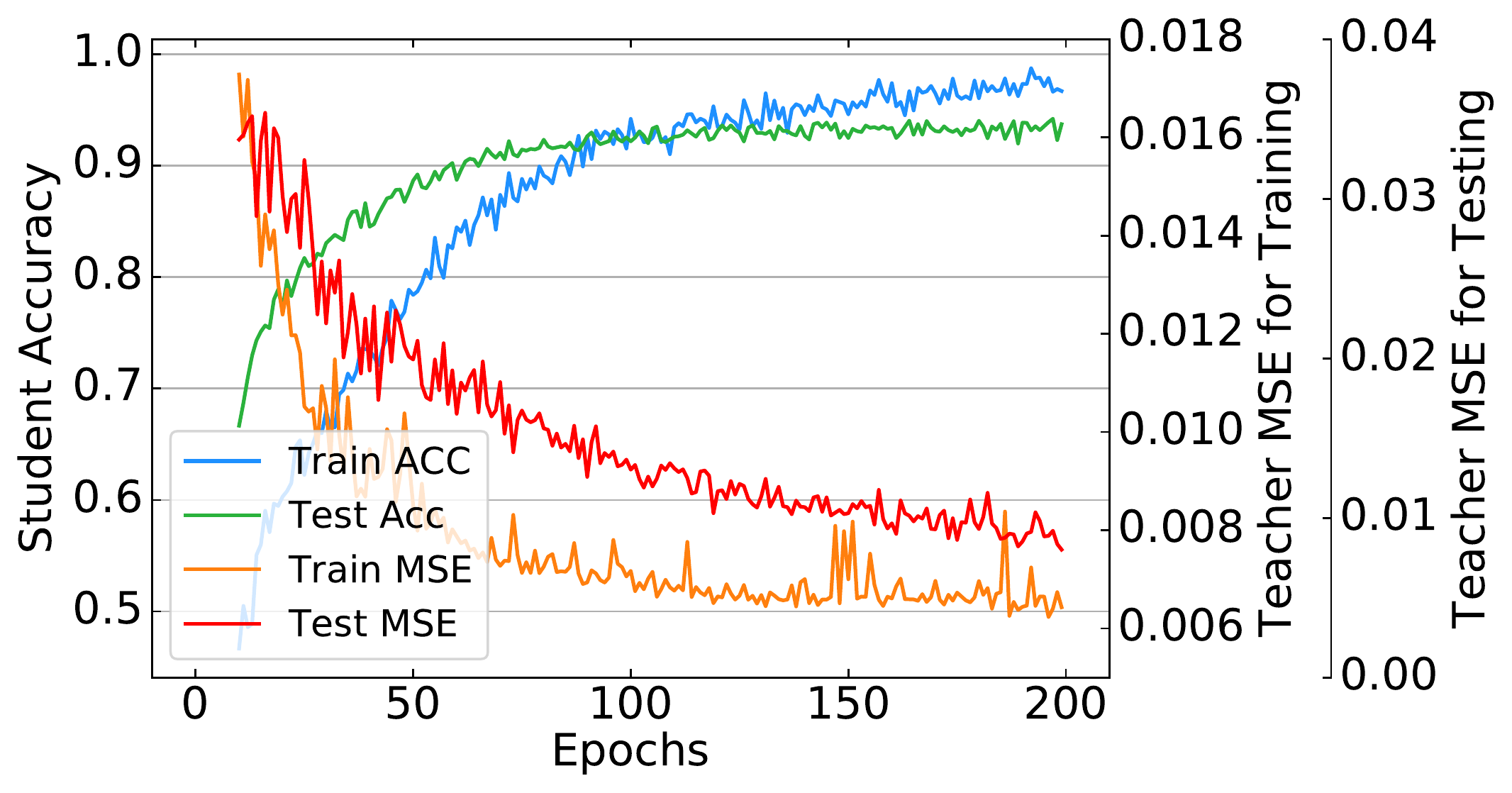} \label{fig:7c}}
		\subfigure[AGSS-TS on CS]{\includegraphics[width=0.238\linewidth]{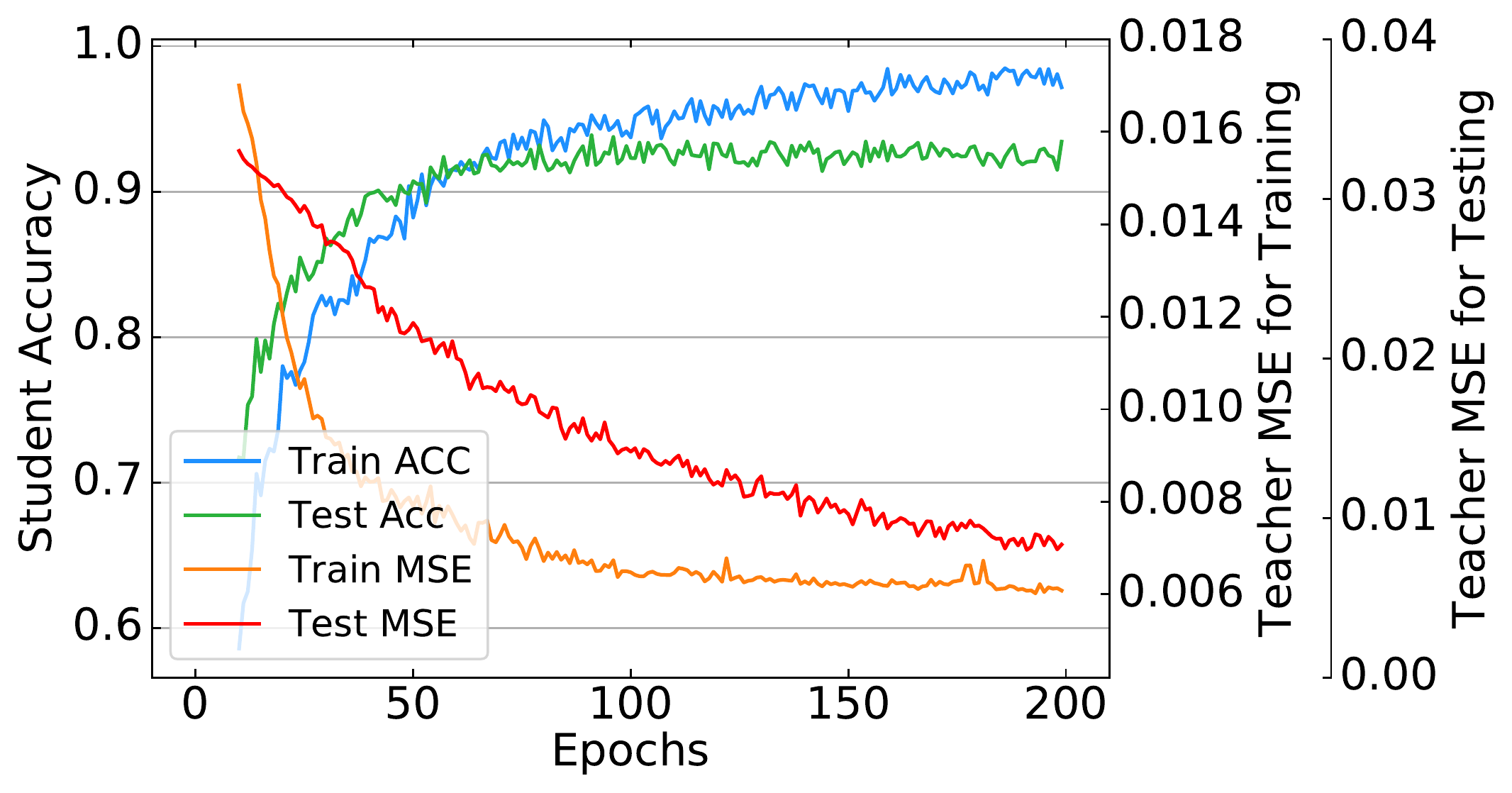} \label{fig:7d}}
	\end{center}
	\vspace{-1em}
	\caption{Learning curves of (1) mean squared errors of teacher probability $\mathbf{p}^{\mathrm{t}}(\mathbf{x})$ over the one-hot labels and (2) classification accuracy, to estimate the quality of the teacher probability $\mathbf{p}^{\mathrm{t}}(\mathbf{x})$.}
	\vspace{-1em}
	\label{fig:7}
\end{figure*}

\textbf{Learning Curves (Q4).} Since the true Bayesian probability $\mathbf{p}^*(\mathbf{x})$ is often unknown in practice, it is not feasible to directly estimate the squared errors between $\mathbf{p}^{\mathrm{t}}(\mathbf{x})$ and $\mathbf{p}^*(\mathbf{x})$. Therefore, we follow \citep{menon2021statistical,zhou2021rethinking}  to estimate the quality of the teacher probability $\mathbf{p}^{\mathrm{t}}(\mathbf{x})$ by computing the Mean Squared Errors (MSE) over one-hot labels. We provide the curves of MSE and classification accuracy during training in Fig.~\ref{fig:7}, from which we observe that the MSE gradually decreases while the classification accuracy gradually increases on both the training and testing sets as the training proceeds. This not only demonstrates the effectiveness of the two proposed adaptive knowledge integration methods but also justifies the proposed theoretical guideline \ref{remark:1}.

\vspace{-0.5em}
\subsection{Evaluation on Knowledge Integration and Teacher Number (Q5)} \label{sec:4.3}
\begin{figure*}[!tbp]
	\begin{center}
		\subfigure[Citeseer]{\includegraphics[width=0.22\linewidth]{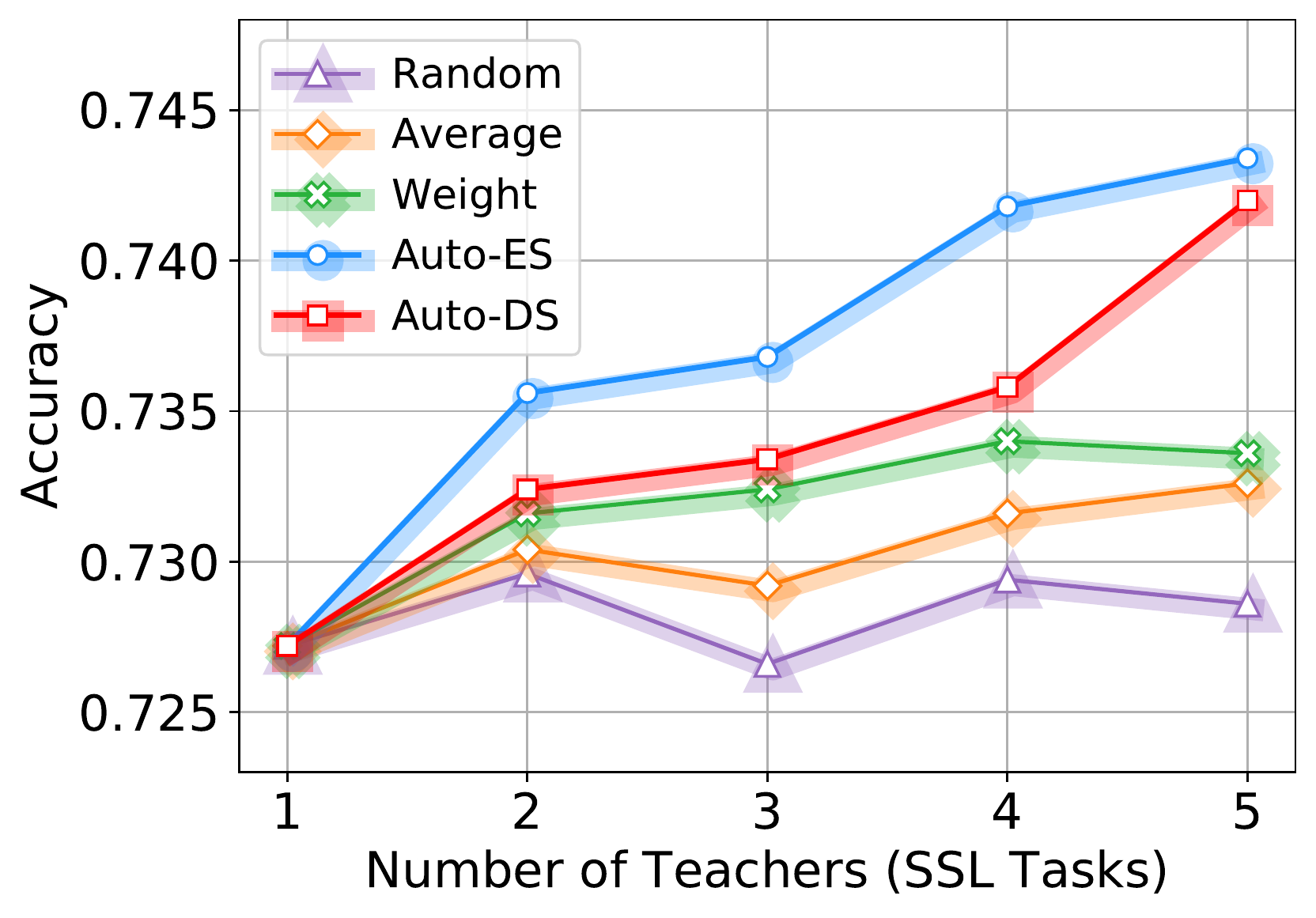} \label{fig:8a}}
		\subfigure[Coauthor-CS]{\includegraphics[width=0.22\linewidth]{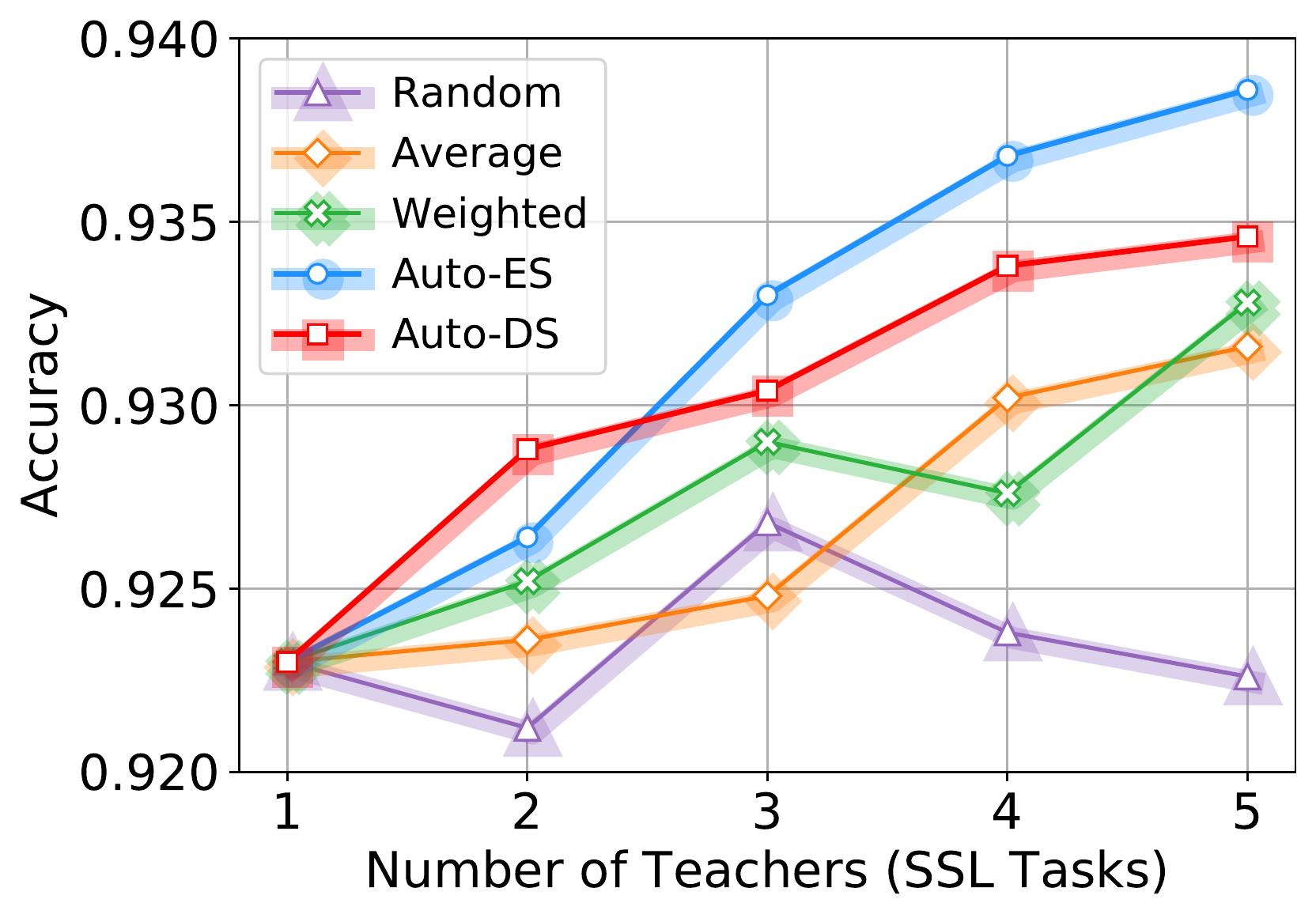} \label{fig:8b}}
		\subfigure[Hyperparameter $\alpha$]{\includegraphics[width=0.254\linewidth]{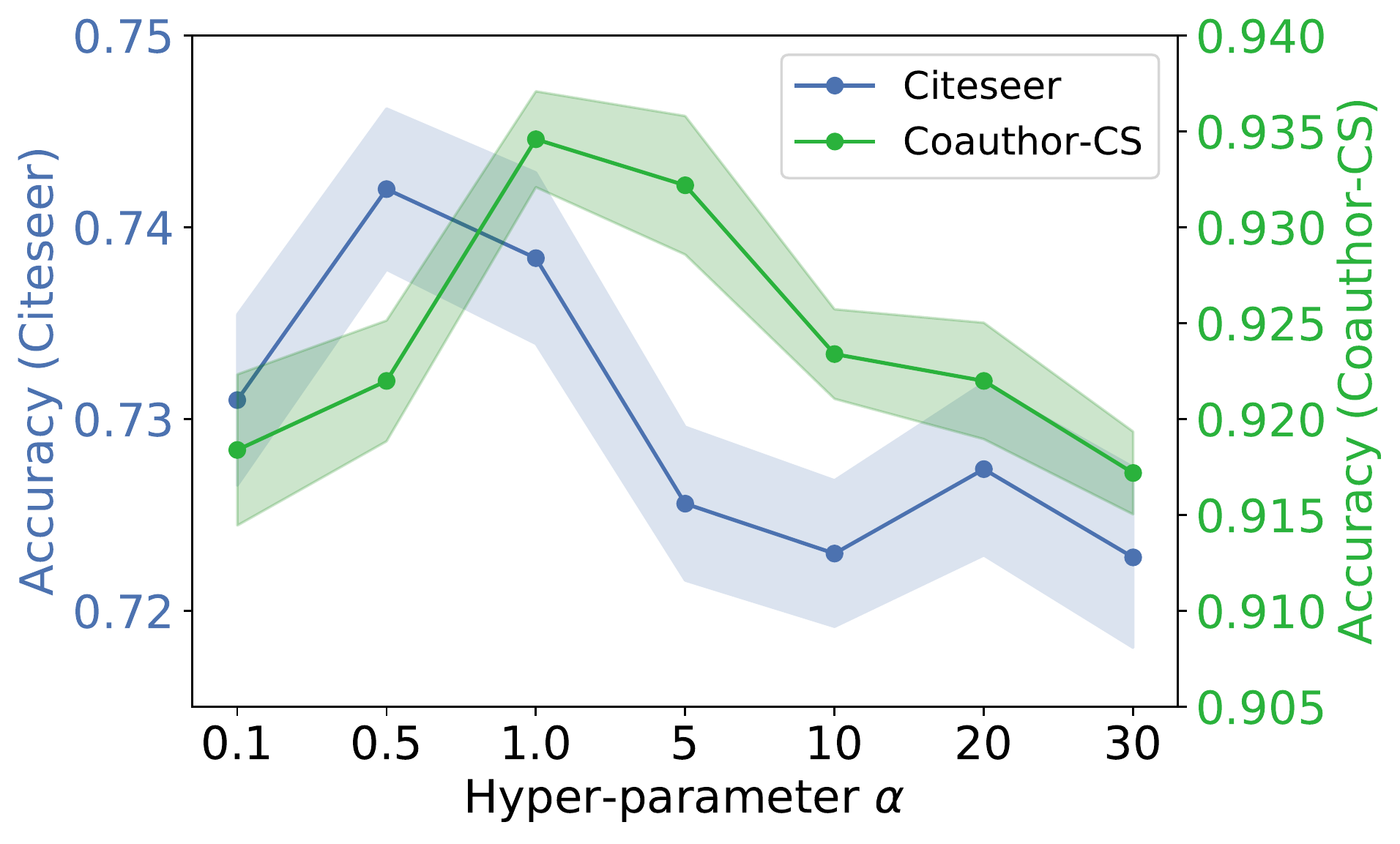} \label{fig:8c}}
		\subfigure[Hyperparameter $\beta$]{\includegraphics[width=0.254\linewidth]{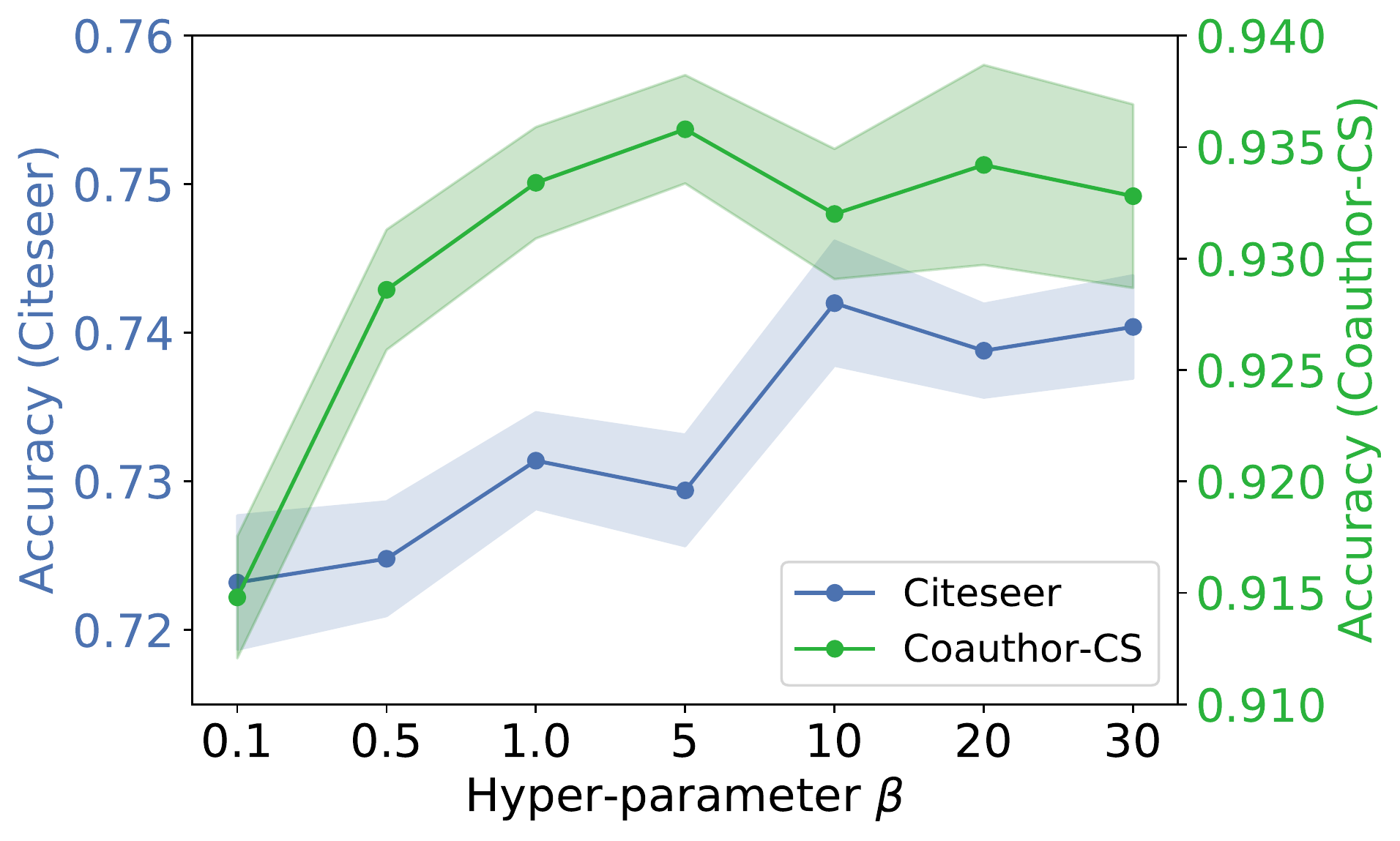} \label{fig:8d}}
	\end{center}
	\vspace{-1em}
	\caption{\textbf{(a-b)} Ablation study on knowledge integration under different number of teachers (with concrete values in \textbf{Appendix I}). \textbf{(c-d)} Parameter sensitivity analyses on loss weights $\alpha$ and $\beta$.}
	\vspace{-1em}
	\label{fig:8}
\end{figure*}

\vspace{-0.5em}
To answer \textbf{Q5}, we compare \texttt{AGSSL-LF} and \texttt{AGSSL-TS} with three heuristic knowledge integration schemes, including (1) \texttt{Random}, setting $\lambda_\gamma(k,i)$ randomly in the range of [0,1]; (2) \texttt{Average}, setting $\lambda_\gamma(k,i)=1/K$ throughout training, and (3) \texttt{Weighted}, calculating cross-entropy as weights on the labeled nodes, and using average weights for unlabeled nodes. For a fair comparison, we perform softmax activation for each scheme to satisfy $\sum_{k=1}^K\lambda_\gamma(k,i)\!=\!1$. We provide the performance of these schemes under five different numbers of teachers in Fig.~\ref{fig:8a} and Fig.~\ref{fig:8b}, from which we can see that (1) \texttt{Random} not only does not benefit from multiple teachers but also is poorer than the one trained with one individual task; (2) \texttt{Average} and \texttt{Weighted} cannot always benefit from multiple teachers; for example, the \texttt{Weighted} scheme trained with five pretext task is inferior to the one trained with four pretext tasks on the Citeseer dataset; (3) \texttt{AGSSL-LF} and \texttt{AGSSL-TS} both perform far better than the other three heuristics under various numbers of teachers. More importantly, both \texttt{AGSSL-LF} and \texttt{AGSSL-TS} can consistently benefit from more teachers, which aligns with Theorem \ref{definition:1} and demonstrates the effectiveness of the two proposed knowledge integration strategies.

\vspace{-0.5em}
\subsection{Evaluation on hyperparameter sensitivity and computational efficiency}
\vspace{-0.5em}
We provide sensitivity analyses for two hyperparameters, loss weights $\alpha$ and $\beta$ in Fig.~\ref{fig:8c} and Fig.~\ref{fig:8d}, from which it is clear that (1) setting the loss weight $\alpha$ of pretext tasks too large or too small is detrimental to extracting informative knowledge; (2) a large $\beta$ usually yields good performance, which illustrates the effectiveness of the distillation term in Eq.~(\ref{equ:4}). In practice, we can determine $\alpha$ and $\beta$ by selecting the model with the highest accuracy on the validation set through the grid search.

Due to space limitation, we place experimental results on \textbf{computational efficiency} in \textbf{Appendix J}, from which we find that compared to jointly training with multiple pretext tasks, \texttt{AGSSL} not only does not increase, but even has advantages in terms of both training time and peak memory usage.

\vspace{-0.5em}
\section{Related Work}
\vspace{-0.5em}
\textbf{Graph Self-supervised Learning (SSL).}
The primary goal of Graph SSL is to learn transferable prior knowledge from abundant unlabeled data through well-designed pretext tasks. There have been up to hundreds of self-supervised pretext tasks proposed in the past few years, and we refer interested readers to the recent surveys \citep{wu2021self,xie2021self,liu2021graph} for more information. Despite the great success, these methods mostly focus on designing more powerful pretext tasks, with little effort to explore how to leverage multiple existing pretext tasks more efficiently.

\textbf{Automated Machine Learning.}
One of the most related topics to us is the automated loss function search \citep{zhao2021autoloss,weber2020automated,hutter2019automated,waring2020automated,yao2018taking}. However, most of these methods are specifically designed for image data and may not be applicable for graph-structured data. For example, the loss function of \texttt{PAIRDIS} involves two nodes, which is hardly compatible with the node-specific loss function of \texttt{CLU}. A recent work JOAO \citep{you2021graph} on graph contrastive learning is proposed to automatically select data augmentation, but it is tailored for graph classification and single-task contrastive learning and is difficult to extend to multi-task self-supervised learning. Another related work is AUX-TS \citep{han2021adaptive}, which adaptively combines different auxiliary tasks in order to generalize to other tasks during the fine-tuning stage of transfer learning, which is hard to extend directly to the self-supervised setting. The closest work to us is AutoSSL \citep{jin2021automated}, which has been discussed in detail in Sec.~\ref{sec:2}.

\vspace{-0.5em}
\section{Conclusion}
\vspace{-0.5em}
Over the past few years, there have been hundreds of self-supervised algorithms proposed, which inspired us to move our attention away from designing more powerful but complex pretext tasks and towards making more effective use of those already on hand. In this paper, we propose a novel multi-teacher knowledge distillation framework for \underline{A}utomated \underline{G}raph \underline{S}elf-\underline{S}upervised \underline{L}earning (\texttt{AGSSL}) to automatically, adaptively, and dynamically learn instance-level self-supervised learning strategies for each node separately. More importantly, we provide a provable theoretical guideline and two adaptive integration strategies to integrate the knowledge from different teachers. While \texttt{AGSSL} automates the selection of pretext tasks for each node, it is still preliminary work, as how to construct a suitable pool of pretext tasks still requires human labor. In this sense, ``full" automation is still desired and needs to be pursued in the future. Meanwhile, despite being developed for Graph SSL, extending \texttt{AGSSL} to other self-supervised tasks for image and text data is also a promising direction.

\clearpage
\bibliography{iclr2023_conference}
\bibliographystyle{iclr2023_conference}

\clearpage
\renewcommand\thefigure{A\arabic{figure}}
\renewcommand\thetable{A\arabic{table}}
\renewcommand\theequation{A.\arabic{equation}}
\setcounter{table}{0}
\setcounter{figure}{0}
\setcounter{theorem}{0}
\setcounter{equation}{0}
\setcounter{definition}{0}

\section*{Appendix}

\subsection*{A. Extensions to the \textit{Pre-train\&Fine-tune}
(\texttt{P\&F}) Setting} 
\subsubsection*{A.1 AutoSSL for the \textit{Pre-train\&Fine-tune} Setting}
To adapt AutoSSL to the \textit{Pre-train\&Fine-tune} setting, the learning objective can be formulated as
\begin{equation}
\theta^{*}, \omega^{*}=\arg \min _{\theta, \omega} \mathcal{L}_{\mathrm{task}}(\theta_{init}, \omega)
\end{equation}
where the initialized parameter $\theta_{init}$ is obtained by optimizing the following objective function
\begin{equation}
\min_{\{\lambda_k\}_{k=1}^K} \mathcal{H}\big(f_{\theta_{init}}(\mathcal{G})\big), \quad \text{s.t.} \text{ } \text{ } \theta_{init}, \{\eta_k^*\}_{k=1}^K=\mathop{\arg\min}_{\theta, \{\eta_k\}_{k=1}^K} \sum_{k=1}^K \lambda_k \mathcal{L}_{\mathrm{ssl}}^{(k)}(\theta,\eta_k)
\end{equation}

\subsubsection*{A.2 Proposed \texttt{AGSSL} for the \textit{Pre-train\&Fine-tune} Setting}
To adapt \texttt{AGSSL} to the \textit{Pre-train\&Fine-tune} setting, the learning objective is defined as
\begin{equation}
\min_{\theta, \omega, \gamma} \mathcal{L}_{\mathrm{task}}\big(\theta, \omega\big)+\beta \frac{\tau^2}{N}\sum_{i=1}^N \mathcal{L}_{KL} \Big(\operatorname{softmax}(\mathbf{z}_i/\tau), \sum_{k=1}^K \lambda_\gamma(k,i) \operatorname{softmax}(\mathbf{h}_i^{(k)}/\tau)\Big)
\end{equation}
where $\mathbf{z}_i = g_\omega(f_\theta(\mathcal{G}, i))$ is the logit of node $v_i$ in the student model and $\mathbf{h}_i^{(k)} = g_{\omega_k^{*}}(f_{\theta_k^{*}}(\mathcal{G}, i))$ is the logit of node $v_i$ in the $k$-th teacher model. $\lambda_\gamma(k,i)$ is a parameterized module that outputs the importance weight of $k$-th pretext task for node $v_i$, which satisfies $\sum_{k=1}^K\lambda_\gamma(k,i)=1$. The parameters $\{\theta_k^{*}\}_{k=1}^K$ and $\{\omega_k^{*}\}_{k=1}^K$ are obtained by optimizing the following objective function
\begin{equation}
\theta^{*}_k, \omega^{*}_k=\arg \min _{(\theta_k, \omega_k)} \mathcal{L}_{\mathrm{task}}(\theta_k^{init}, \omega_k), \quad \text{s.t.} \text{ } \text{ } \theta_k^{init}, \eta_k^*=\mathop{\arg\min}_{\theta_k, \eta_k} \mathcal{L}_{\mathrm{ssl}}^{(k)}(\theta_k, \eta_k)
\end{equation}

\subsection*{B. Distillation Objective Rewriting}
Given $\widetilde{\mathbf{z}}_i=\operatorname{softmax}(\mathbf{z}_i/\tau)$, $\widetilde{\mathbf{h}}_i^{(k)}=\operatorname{softmax}(\mathbf{h}_i^{(k)}/\tau)$, and $\mathbf{p}^{\mathrm{t}}(\mathbf{x}_i)=\sum_{k=1}^K \lambda_\gamma(k,i) \widetilde{\mathbf{h}}_i^{(k)}$, we derive how to rewrite the second term of Eq.~(\ref{equ:4}), the \textit{\textbf{distillation objective}}, in the form of $\widetilde{R}(\theta,\omega)$ in Eq.~(\ref{equ:5}),
\begin{equation}
\begin{aligned}
&\frac{1}{N}\sum_{i=1}^N \mathcal{L}_{KL} \Big(\operatorname{softmax}(\mathbf{z}_i/\tau), \sum_{k=1}^K \lambda_\gamma(k,i) \operatorname{softmax}(\mathbf{h}_i^{(k)}/\tau)\Big) \\
& =\frac{1}{N}\sum_{i=1}^N \mathcal{L}_{KL} \Big(\widetilde{\mathbf{z}}_i, \sum_{k=1}^K \lambda_\gamma(k,i) \widetilde{\mathbf{h}}_i^{(k)}\Big) =\frac{1}{N}\sum_{i=1}^N \mathcal{L}_{KL} \Big(\widetilde{\mathbf{z}}_i, \mathbf{p}^{\mathrm{t}}(\mathbf{x}_i)\Big)\\
& = \frac{1}{N}\sum_{i=1}^N \mathbf{p}^{\mathrm{t}}(\mathbf{x}_i)\log\frac{\mathbf{p}^{\mathrm{t}}(\mathbf{x}_i)}{\widetilde{\mathbf{z}}_i} = \frac{1}{N}\sum_{i=1}^N \mathcal{I}\big(\mathbf{p}^{\mathrm{t}}(\mathbf{x}_i)\big) - \mathbf{p}^{\mathrm{t}}(\mathbf{x}_i) \log \widetilde{\mathbf{z}}_i
\end{aligned}
\end{equation}
where $\mathcal{I}(\cdot)$ denotes the information entropy, and since $\mathbf{p}^{\mathrm{t}}(\mathbf{x}_i)$ is fixed during training, we can directly omit the term $\mathcal{L}\big(\mathbf{p}^{\mathrm{t}}(\mathbf{x}_i)\big)$ and derive the following proportional equation, as follows
\begin{equation*}
\hspace{-0.5em}
\begin{aligned}
\frac{1}{N}\sum_{i=1}^N \mathcal{L}\big(\mathbf{p}^{\mathrm{t}}(\mathbf{x}_i)\big) \!-\! \mathbf{p}^{\mathrm{t}}(\mathbf{x}_i) \log \widetilde{\mathbf{z}}_i  \propto \frac{1}{N}\sum_{i=1}^N - \mathbf{p}^{\mathrm{t}}(\mathbf{x}_i) \log \widetilde{\mathbf{z}}_i \!=\! \frac{1}{N}\sum_{i=1}^N \mathbf{p}^{\mathrm{t}}(\mathbf{x}_i)\mathbf{l}\big(g_\omega(f_\theta(\mathbf{x}_i))\big)\doteq \widetilde{R}(\theta,\omega)
\end{aligned}
\end{equation*}
where $\mathbf{l}\big(g_\omega(f_\theta(\mathbf{x}_i))\big)\!=\!\big(\ell(1, \widetilde{\mathbf{z}}_i),\ell(2, \widetilde{\mathbf{z}}_i),\cdots,\ell(C, \widetilde{\mathbf{z}}_i)\big)$ denotes the cross-entropy loss vector, and we have $\widetilde{\mathbf{z}}_i=\operatorname{softmax}(\mathbf{z}_i/\tau)=\operatorname{softmax}(g_\omega(f_\theta(\mathbf{x}_i))/\tau)$.

\subsection*{C. Proof on Proposition \ref{theorem:1}}
\begin{theorem}
Consider an interagted teacher  $\mathbf{p}^{\mathrm{t}}(\mathbf{x})$ and a Bayesian teacher $\mathbf{p}^*(\mathbf{x})$. For any GNN encoder $f_\theta(\cdot)$ and prediction head $g_\omega(\cdot)$, the difference between the distillation objective $\widetilde{R}(\theta,\omega)$ and Bayesian objective $R(\theta,\omega)$ is bounded by the Mean Square Error (MSE) of their probabilities,
\begin{equation}
\mathbb{E}\left[\left(\widetilde{R}(\theta,\omega)-R(\theta,\omega)\right)^{2}\right] \leq \frac{1}{N} \mathbb{V}\Big[\mathbf{p}^{\mathrm{t}}(\mathbf{x})^{\mathrm{t}} \mathbf{l}\big(g_\omega(f_\theta(\mathbf{x}))\big)\Big]+\mathcal{O}\Big(\mathbb{E}\big[\|\mathbf{p}^{\mathrm{t}}(x)-\mathbf{p}^{*}(x)\|_2\big]\Big)^2
\end{equation}
\end{theorem}

\begin{proof}
Let us start the derivation from the left side of the equation, as follows
\begin{equation}
\mathbb{E}\left[\left(\widetilde{R}(\theta,\omega)-R(\theta,\omega)\right)^{2}\right]=\mathbb{V}\left[\left(\widetilde{R}(\theta,\omega)-R(\theta,\omega)\right)\right]+\mathbb{E}\left[\left(\widetilde{R}(\theta,\omega)-R(\theta,\omega)\right)\right]^{2}
\label{equ:A7}
\end{equation}
Since $R(\theta,\omega)$ can be considered as a constant in practice, we have
\begin{equation}
\begin{aligned}
\mathbb{V}\left[\left(\widetilde{R}(\theta,\omega)-R(\theta,\omega)\right)\right] & = \mathbb{V}\left[\widetilde{R}(\theta,\omega)\right] + \mathbb{V}\left[R(\theta,\omega)\right] - Cov\left(\widetilde{R}(\theta,\omega), R(\theta,\omega)\right) \\
& = \mathbb{V}\left[\widetilde{R}(\theta,\omega)\right] = \frac{1}{N} \mathbb{V}\Big[\mathbf{p}^{\mathrm{t}}(\mathbf{x})^{\mathrm{t}} \mathbf{l}\big(g_\omega(f_\theta(\mathbf{x}))\big)\Big]
\end{aligned}
\label{equ:A8}
\end{equation}
Next, we consider the second term on the right-hand side in Eq.~(\ref{equ:A7}), as follows
\begin{equation}
\begin{aligned}
\mathbb{E}\left[\left(\widetilde{R}(\theta,\omega)-R(\theta,\omega)\right)\right]^{2} 
& = \mathbb{E}_{\mathbf{x}}\Big[\mathbf{p}^{\mathrm{t}}(\mathbf{x})^{\mathrm{t}} \mathbf{l}\big(g_\omega(f_\theta(\mathbf{x}))\big)-\mathbf{p}^{*}(\mathbf{x})^{\mathrm{t}} \mathbf{l}\big(g_\omega(f_\theta(\mathbf{x}))\big)\Big]^{2} \\
& = \mathbb{E}_{\mathbf{x}}\Big[\big(\mathbf{p}^{\mathrm{t}}(\mathbf{x})^{\mathrm{t}} - \mathbf{p}^{*}(\mathbf{x})^{\mathrm{t}}\big) \mathbf{l}\big(g_\omega(f_\theta(\mathbf{x}))\big)\Big]^{2} \\
& \leq \mathbb{E}_{\mathbf{x}}\Big[\big\|\mathbf{p}^{\mathrm{t}}(\mathbf{x})^{\mathrm{t}} - \mathbf{p}^{*}(\mathbf{x})^{\mathrm{t}}\big\|_2 \cdot \big\| \mathbf{l}\big(g_\omega(f_\theta(\mathbf{x}))\big)\big\|_2\Big]^{2} \\
& \doteq \mathcal{O}\Big(\mathbb{E}\big[\|\mathbf{p}^{\mathrm{t}}(x)-\mathbf{p}^{*}(x)\|_2\big]\Big)^2
\end{aligned}
\label{equ:A9}
\end{equation}
where the inequality holds according to Cauchy-Schwartz inequality \citep{steele2004cauchy}. Combining the derivations of Eq.~(\ref{equ:A8}) and Eq.~(\ref{equ:A9}) into Eq.~(\ref{equ:A7}), we obtain the final inequality as follows
\begin{equation}
\mathbb{E}\left[\left(\widetilde{R}(\theta,\omega)-R(\theta,\omega)\right)^{2}\right] \leq \frac{1}{N} \mathbb{V}\Big[\mathbf{p}^{\mathrm{t}}(\mathbf{x})^{\mathrm{t}} \mathbf{l}\big(g_\omega(f_\theta(\mathbf{x}))\big)\Big]+\mathcal{O}\Big(\mathbb{E}\big[\|\mathbf{p}^{\mathrm{t}}(x)-\mathbf{p}^{*}(x)\|_2\big]\Big)^2
\end{equation}
\end{proof}

\subsection*{D. Proof on Theorem \ref{definition:1}}
\begin{definition} \label{definition:1}
Define $\Delta(K) \!=\! \min\big\|\mathbf{p}^{\mathrm{t}}(\mathbf{x}_i)-\mathbf{p}^{*}(\mathbf{x}_i)\big\|_2 \!=\! \min\big\|\sum_{k=1}^K \lambda_\gamma(k,i)\widetilde{\mathbf{h}_i}^{(k)}\!-\!\mathbf{p}^{*}(\mathbf{x}_i)\big\|_2$ with $K (K \geq 1)$ given teachers, then we have (1) $\Delta(K+1) \leq \Delta(K)$, and (2) $\lim _{K \rightarrow \infty} \Delta(K)=0$.
\end{definition}

\textbf{Proof.}
Let us simplify the symbol $\lambda_\gamma(k,i)$ to $\lambda_k$ and consider the case with $K$ teachers, we have
\begin{equation}
\begin{aligned}
    \{\lambda_k^*\}_{k=1}^K = & \arg\min_{\{\lambda_k\}_{k=1}^K}\big\|\sum_{k=1}^K \lambda_k\widetilde{\mathbf{h}}_i^{(k)}-\mathbf{p}^{*}(x)\big\| \\ \Delta \mathbf{p}_K =& \sum_{k=1}^K \lambda_k^*\widetilde{\mathbf{h}}_i^{(k)}-\mathbf{p}^{*}(x), \Delta(K) =  \|\Delta \mathbf{p}_K\|_2
\end{aligned}
\end{equation}
Next, let's consider the case with $(K+1)$ teachers, as follows
\begin{equation}
\begin{aligned}
    \Delta(K+1) =  & \min_{\{\lambda_k\}_{k=1}^K}\big\|\sum_{k=1}^{K+1} \lambda_k\widetilde{\mathbf{h}_i}^{(k)}-\mathbf{p}^{*}(\mathbf{x}_i)\big\|_2  \\ 
    \leq & \min_{\lambda_{K+1}}\big\|\sum_{k=1}^{K} \lambda_k^*\widetilde{\mathbf{h}_i}^{(k)} + \lambda_{K+1}\widetilde{\mathbf{h}_i}^{(K+1)}-\mathbf{p}^{*}(\mathbf{x}_i)\big\|_2 \\
    = & \min_{\lambda_{K+1}}\big\|\lambda_{K+1}\widetilde{\mathbf{h}_i}^{(K+1)}+\Delta\mathbf{p}_K\big\|_2 \\
    \leq & \sin\left(arc\cos\frac{<\Delta \mathbf{p}_{K}, \widetilde{\mathbf{h}_i}^{(K+1)}>}{\big\|\Delta \mathbf{p}_{K}\big\|_2\cdot\big\|\widetilde{\mathbf{h}_i}^{(K+1)}\big\|_2}\right) \cdot \big\|\Delta\mathbf{p}_K\big\|_2 \\
    \leq & \big\|\Delta\mathbf{p}_K\big\|_2 = \Delta(K)
\end{aligned}
\label{equ:A12}
\end{equation}

where the equality in the fourth row of Eq.~(\ref{equ:A12}) holds under the condition that 
\begin{equation}
\lambda_{k+1}=-\frac{<\Delta \mathbf{p}_K, \widetilde{\mathbf{h}_i}^{(K+1)}>}{\big\|\Delta \mathbf{p}_K\big\|_2\cdot\big\|\widetilde{\mathbf{h}_i}^{(K+1)}\big\|_2} \cdot \frac{\big\|\Delta \mathbf{p}_K\big\|_2}{\big\|\widetilde{\mathbf{h}_i}^{(K+1)}\big\|_2} = -\frac{<\Delta \mathbf{p}_K, \widetilde{\mathbf{h}_i}^{(K+1)}>}{\big\|\widetilde{\mathbf{h}_i}^{(K+1)}\big\|^2_2}    
\end{equation}

Let $K\geq 2$ be the number of teachers, and the results of the $K$-th iteration can be defined as follows:
\begin{equation*}
\begin{small}
\begin{aligned}
    \Delta(K) \leq & \sin\left(arc\cos\frac{<\Delta \mathbf{p}_{K-1}, \widetilde{\mathbf{h}_i}^{(K)}>}{\big\|\Delta \mathbf{p}_{K-1}\big\|_2\cdot\big\|\widetilde{\mathbf{h}_i}^{(K)}\big\|_2}\right) \cdot \Delta(K-1) \\
    \leq & = \sin\left(arc\cos\frac{<\Delta \mathbf{p}_{K-1}, \widetilde{\mathbf{h}_i}^{(K)}>}{\big\|\Delta \mathbf{p}_{K-1}\big\|_2\cdot\big\|\widetilde{\mathbf{h}_i}^{(K)}\big\|_2}\right) \cdot \sin\left(arc\cos\frac{<\Delta \mathbf{p}_{K-2} \widetilde{\mathbf{h}_i}^{(K-1)}>}{\big\|\Delta \mathbf{p}_{K-2}\big\|_2\cdot\big\|\widetilde{\mathbf{h}_i}^{(K-1)}\big\|_2}\right) \cdot \Delta(K-2) \\
    \leq & \cdots \\
    \leq & \prod_{k=2}^{K}\sin\left(arc\cos\frac{<\Delta \mathbf{p}_{k-1}, \widetilde{\mathbf{h}_i}^{(k)}>}{\big\|\Delta \mathbf{p}_{k-1}\big\|_2\cdot\big\|\widetilde{\mathbf{h}_i}^{(k)}\big\|_2}\right) \cdot \Delta(1)
\end{aligned}
\end{small}
\end{equation*}

Since $\sin\left(arc\cos\frac{<\Delta \mathbf{p}_{k-1}, \widetilde{\mathbf{h}_i}^{(k)}>}{\big\|\Delta \mathbf{p}_{k-1}\big\|_2\cdot\big\|\widetilde{\mathbf{h}_i}^{(k)}\big\|_2}\right) \leq 1$, and the equality holds when and only when $\Delta \mathbf{p}_{k-1}$ and $\widetilde{\mathbf{h}_i}^{(k)}$ are orthogonal, which in practice is hard to be satisfied, we have $\lim _{K \rightarrow \infty} \Delta(K)=0$.

\subsection*{E. Pseudo Code of \texttt{AGSSL}}
The pseudo-code of the proposed \texttt{AGSSL} framework is summarized in Algorithm~\ref{algo:1}.

\begin{algorithm}[!htbp]
	\caption{Algorithm for the \textit{Multi-teacher Knowledge Distillation} framework for \texttt{AGSSL}}
	\label{algo:1}
	\begin{algorithmic}[1]
		\Require Graph $\mathcal{G}=(\mathcal{V},\mathcal{E},\mathbf{X})$, Number of Pretext Tasks: $K$, and Number of Epochs: $T$. 
		
		\Ensure Predicted Labels $\mathcal{Y}_U$, GNN Enocder $f_\theta(\cdot)$, and Prediction Head $g_\omega(\cdot)$.
		
		\State Randomly initialize the parameters of $K$ teacher models and a student model.
		\State Pre-train each teacher with one pretext task to get pre-trained parameters $\{\theta_k^*, \omega_k^*\}_{k=1}^K$.
		\For{$t$ $\in$ \{0, 1, $\cdots$, $T-1$\}}
		\State Output logits $\big\{\mathbf{h}_i^{(k)} = g_{\omega_k^*}(f_{\theta_k^*}(\mathcal{G}, i))\big\}_{k=1}^K$ from the pre-trained teachers and freeze them.
		\vspace{-0.2em}
		\State Integrate the knowledge of different teachers by $\mathbf{p}^{\mathrm{t}}(\mathbf{x}_i)=\sum_{k=1}^K \lambda_\gamma(k,i) \operatorname{softmax}(\mathbf{h}_i^{(k)}/\tau)$.
		\State Jointly perform distillation by Eq.~(\ref{equ:4}) and optimize the function $\lambda_\gamma(\cdot,\cdot)$ with loss $\mathcal{L}_W$.
		
		\EndFor
		\State \textbf{return} Predicted labels $\mathcal{Y}_U$, GNN encoder $f_\theta(\cdot)$, and prediction head $g_\omega(\cdot)$.
	\end{algorithmic}
\end{algorithm}

\subsection*{F. Dataset Statistics} 
\emph{Eight} publicly available graph datasets are used to evaluate the proposed \texttt{AGSSL} framework. An overview summary of the statistical characteristics of datasets is given in Tab.~\ref{tab:A1}. For the three small-scale datasets, namely Cora, Citeseer, and Pubmed, we follow the data splitting strategy in \citep{kipf2016semi}. For the four large-scale datasets, namely Coauthor-CS, Coauthor-Physics, Amazon-Photo, and Amazon-Computers, we follow \citep{zhang2021graph,luo2021distilling} to randomly split the data into train/val/test sets, and each random seed corresponds to a different splitting. For the ogbn-arxiv dataset, we use the public data splits provided by the authors \citep{hu2020open}.

\begin{table*}[!htbp]
\begin{center}
\caption{Statistical information of the datasets.}
\label{tab:A1}
\resizebox{\textwidth}{!}{
\begin{tabular}{lccccccccc}

\toprule
\textbf{Dataset} & \texttt{Cora} & \texttt{Citeseer} & \texttt{Pubmed} & \texttt{Photo} & \texttt{CS} & \texttt{Physics} & \texttt{Computers} & \texttt{ogbn-arxiv} \\ \midrule
\textbf{$\#$ Nodes} & 2708 & 3327 & 19717 & 7650 & 18333 & 34493 & 13752 & 169343 \\
\textbf{$\#$ Edges} & 5278 & 4614 & 44324 & 119081 & 81894 & 247962 & 245861 & 1166243 \\
\textbf{$\#$ Features} & 1433 & 3703 & 500 & 745 & 6805 & 8415 & 767 & 128 \\
\textbf{$\#$ Classes} & 7 & 6 & 3 & 8 & 15 & 5 & 10 & 40 \\
\textbf{Label Rate} & 5.2\% & 3.6\% & 0.3\% & 2.1\% & 1.6\% & 0.3\% & 1.5\% & 53.7\% \\ \bottomrule

\end{tabular}} \vspace{-1em}
\end{center}
\end{table*}

\subsection*{G. Hyperparameter Settings}
The following hyperparameters are set the same for all datasets: Adam optimizer with learning rate $lr$ = 0.01 (0.001 for \texttt{ogb-arxiv}) and weight decay $w$ = 5e-4; Epoch $E$ = 500; Layer number $L$ = 1 (2 for \texttt{ogb-arxiv}). The other dataset-specific hyperparameters are determined by an AutoML toolkit NNI with the hyperparameter search spaces as: hidden dimension $F=\{32, 64, 128, 256, 512\}$; distillation temperature $\tau=\{1, 1.2, 1.5, 2, 3, 4, 5\}$, and loss weights $\alpha,\beta=\{0.1, 0.5, 1, 5, 10, 20, 30\}$. For a fairer comparison, the model with the highest validation accuracy is selected for testing. Besides, the best hyperparameter choices for each dataset are available in the supplementary.  Moreover, the experiments on both baselines and our approach are implemented based on the standard implementation in the DGL library \citep{wang2019dgl} using the PyTorch 1.6.0 with Intel(R) Xeon(R) Gold 6240R @ 2.40GHz CPU and NVIDIA V100 GPU.

\subsection*{H. Details on Five Pretext Tasks}
In this paper, we evaluate the capability of \texttt{AGSSL} in automatic pretext tasks combinatorial search with five classical pretext tasks, including PAR \citep{you2020does}, CLU \citep{you2020does}, DGI \citep{velickovic2019deep}, PAIRDIS \citep{jin2020self}, and PAIRSIM \citep{jin2020self}. Our motivations for selecting these five pretext tasks are 4-fold: (1) \emph{Fair comparison.} To make a fair comparison with previous methods (e.g., AutoSSL), we keep in line with it in the setting of pretext tasks, i.e., using the same combination of tasks. (2) \emph{Simple but classical.} We should pick those pretext tasks that are simple but classical enough, rather than those that are overly complex, not time-tested, and not well known. This is to avoid, whether the resulting performance gains come from our proposed \texttt{AGSSL} or from the complexity of the selected pretext task itself, becoming incomprehensible and hard to explain. (3) \emph{Comprehensive.} Different pretext tasks implicitly involve different inductive biases, so it is important to consider different aspects comprehensively when selecting pretext tasks, rather than picking too many homogeneous and similar tasks. (4) \emph{Applicability.} There is no conflict at all between Graph SSL automation and designing more powerful pretext tasks; as a general framework, \texttt{AGSSL} is applicable to other more complex self-supervised tasks. However, the focus of this paper is on the knowledge distillation framework rather than on the specific task design, and it is also impractical to enumerate and compare all existing graph SSL methods in a limited space.

\textbf{PAR and CLU.} The pretext task of Node Clustering (CLU) pre-assigns a pseudo-label $\widehat{y}_i$, e.g., the cluster index, to each node $v_i\in\mathcal{V}$ by $K$-means clustering algorithm \citep{macqueen1965some}. The learning objective of this pretext task is then formulated as a classification problem, as follows
\begin{equation}
\mathcal{L}_{\mathrm{ssl}}\left(\theta, \eta\right)=\frac{1}{N}\sum_{v_i \in \mathcal{V}} \ell\Big(g_{\eta}(f_{\theta}(\mathcal{G}, i), \widehat{y}_{i}\Big)
\end{equation}
When node attributes are not available, another choice to obtain pseudo-labels is based on the topology of the graph structure. Specifically, the pretext task of graph partitioning (PAR) predicts partition pseudo-labels obtained by the Metis graph partition \citep{karypis1998fast}. The two tasks, CLU and PAR, are very similar, but they extract \textbf{\emph{feature-level}} and \textbf{\emph{topology-level}} knowledge from the graph, respectively. A key hyperparameter of them is the category number of pseudo-labels \#P, which is set to \#P=10 for CLU and \#P=400 (100 for Amazon-Photo and Amazon-Computers, 1000 for Citeseer) for PAR, following the settings in \citep{jin2021automated}. In practice, CLU can be easily extended to other variants by adopting other clustering methods \citep{wu2022deep,wu2022generalized}. 

\textbf{DGI.} Deep Graph Infomax (DGI) is proposed to contrast the node representations and corresponding high-level summary of graphs. First, it applies an augmentation transformation $\mathcal{T}(\cdot)$ to obtain an augmented graph $\widetilde{\mathcal{G}}= \mathcal{T}(\mathcal{G})$. Then a shared graph encoder $f_{\theta}(\cdot)$ is applied to obtain node embeddings $\mathbf{h}_i=f_{\theta}(\mathcal{G}, i)$ and $\widetilde{\mathbf{h}}_i=f_{\theta}(\widetilde{\mathcal{G}}, i)$. Beside, a global mean pooling is applied to obtain the graph-level representation $\mathbf{h}_{\widetilde{g}}=\frac{1}{N}\sum_{i=1}^N \widetilde{\mathbf{h}}_i$. Finally, the learning objective is defined as follows
\begin{equation}
\mathcal{L}_{\mathrm{ssl}}(\theta) = - \frac{1}{N} \sum_{v_i\in\mathcal{V}}\mathcal{MI}\left(\mathbf{h}_{\widetilde{g}}, \mathbf{h}_i\right)
\end{equation}
where $\mathcal{MI}(\cdot,\cdot)$ is the InfoNCE mutual information estimator \citep{gutmann2010noise}, where the negative samples to contrast with $\mathbf{h}_{\widetilde{g}}$ is $\{\mathbf{h}_{j}\}_{j\neq i}$. The pretext task of DGI extracts knowledge at the graph level. To improve the computational efficiency for large-scale graphs, we will randomly sample 2000 nodes to contrast the representations between these sampled nodes and the whole graph.

\textbf{PAIRDIS.} The pretext task of PAIRDIS aims to guide the model to preserve \textbf{\emph{global topology information}} by predicting the shortest path length between nodes. It first randomly samples a certain amount of node pairs $\mathcal{S}$ and calculates the pairwise node shortest path length $d_{i,j}=d(v_i,v_j)$ for node pairs $(v_i,v_j)\in\mathcal{S}$. Furthermore, it groups the shortest path lengths into four categories: $C_{i,j}\!=\!0, C_{i,j}\!=\!1, C_{i,j}\!=\!2$, and $C_{i,j}\!=\!3$ corresponding to $d_{i,j}\!=\!1, d_{i,j}\!=\!2, d_{i,j}\!=\!3$, and $d_{i,j}\geq4$, respectively. The learning objective can be formulated as a multi-class classification problem, 
\begin{equation}
\mathcal{L}_{\mathrm{ssl}}\left(\theta, \eta\right)=\frac{1}{|\mathcal{S}|}\sum_{(v_{i}, v_{i}) \in \mathcal{S}} \ell\Big(g_{\eta}\big(|f_{\theta}(\mathcal{G})_{v_{i}}-f_{\theta}(\mathcal{G})_{v_{j}}|\big), C_{i,j}\Big) 
\end{equation}
where $\ell(\cdot)$ denotes the cross entropy loss and $g_{\eta}(\cdot)$ linearly maps the input to a 1-dimension value. A key hyperparameter in PAIRDIS is the size of $\mathcal{S}$, which is set to $|\mathcal{S}|=400$ for all eight datasets.

\textbf{PAIRSIM}. Unlike PAIRDIS, which focuses on the global topology, PAIRSIM adopts link prediction as a pretext task to predict feature similarities between node pairs and thus capture \textbf{\emph{local connectivity information}} from the graph. PAIRSIM first masks $m$ edges $\mathcal{M} \in \mathcal{E}$ and also samples $m$ edges $\overline{\mathcal{M}} \in \{(v_i,v_j)|v_i,v_j\in\mathcal{V}$ and $(v_i,v_j)\notin\mathcal{E}\}$. Then, the learning objective of PAIRSIM is to predict whether there exists a link between a given node pair, which can be formulated as follows
\begin{equation*}
\mathcal{L}_{\mathrm{ssl}}\left(\theta, \eta\right)  = \frac{1}{2m}\Big( \sum_{e_{i,j} \in \mathcal{M}} \ell\big(g_{\eta}(|f_{\theta}(\mathcal{G},i)-f_{\theta}(\mathcal{G},j)|), 1\big)+  \sum_{e_{i,j} \in \overline{\mathcal{M}}}  \ell\big(g_{\eta}(|f_{\theta}(\mathcal{G},i)-f_{\theta}(\mathcal{G},j)|), 0\big)\Big)
\end{equation*}
where $\ell(\cdot)$ denotes the cross entropy loss and $g_{\eta}(\cdot)$ linearly maps the input to a 1-dimension value. The pretext task of PAIRSIM aims to help GNN learn more local structural information. A key hyperparameter in PAIRSIM is the size of $\mathcal{M}$, which is set to $|\mathcal{M}|=400$ for all eight datasets.

\begin{table*}[!htbp]
\begin{center}
\vspace{-1em}
\caption{Ablation study on knowledge integration under different number of teachers, where \textbf{bold} and \underline{underline} denote the best and second metrics for each teacher number, respectively. The best performance (i.e., the optimal teacher number) for each integration shceme is marked in \textcolor{blue}{blue}.}
\label{tab:A2}
\resizebox{\textwidth}{!}{
\begin{tabular}{lccccc|ccccc}

\toprule
\multirow{2}{*}{\textbf{Method}} & \multicolumn{5}{c|}{\texttt{Citeseer}}                                    & \multicolumn{5}{c}{\texttt{Coauthor-CS}}                                  \\ \cmidrule(l){2-11} 
                        & 1 (+\texttt{PAR})    & 2 (+\texttt{CLU})    & 3 (+\texttt{DGI})    & 4 (+\texttt{PAIRDIS}) & 5 (+\texttt{PAIRSIM}) & 1 (+\texttt{PAR})    & 2 (+\texttt{CLU})    & 3 (+\texttt{DGI})    & 4 (+\texttt{PAIRDIS}) & 5 (+\texttt{PAIRSIM}) \\ \midrule
\texttt{Random}                  & $72.72_{\pm0.36}$ & \textcolor{blue}{$72.96_{\pm0.47}$} & $72.66_{\pm0.39}$ & $72.94_{\pm0.43 }$ & $72.86_{\pm0.45 }$ & $92.30_{\pm0.67}$ & $92.12_{\pm0.54}$ & \textcolor{blue}{$92.68_{\pm0.47}$} & $92.38_{\pm0.53 }$ & $92.26_{\pm0.64}$  \\
\texttt{Average}                & $72.72_{\pm0.36}$ & $73.04_{\pm0.34}$ & $72.92_{\pm0.42}$ & $73.16_{\pm0.39 }$ & \textcolor{blue}{$73.26_{\pm0.37 }$} & $92.30_{\pm0.67}$ & $92.36_{\pm0.49}$ & $92.48_{\pm0.60}$ & $93.02_{\pm0.55 }$ & \textcolor{blue}{$93.16_{\pm0.47}$}  \\
\texttt{Weighted}                 & $72.72_{\pm0.36}$ & $73.16_{\pm0.32}$ & $73.24_{\pm0.46}$ & \textcolor{blue}{$73.40_{\pm0.43 }$} & $73.36_{\pm0.40 }$ & $92.30_{\pm0.67}$ & $92.52_{\pm0.46}$ & $92.90_{\pm0.51}$ & $92.76_{\pm0.39 }$ & \textcolor{blue}{$93.28_{\pm0.53}$}  \\
\texttt{AGSSL-LF}             & \cellcolor{gray!20}$72.72_{\pm0.36}$ & \cellcolor{gray!20}$\textbf{73.56}_{\pm0.39}$ & \cellcolor{gray!20}$\textbf{73.68}_{\pm0.33}$ & \cellcolor{gray!20}$\textbf{74.18}_{\pm0.40 }$ & \textcolor{blue}{\cellcolor{gray!20}$\textbf{74.34}_{\pm0.31 }$} & \cellcolor{gray!20}$92.30_{\pm0.67}$ & \cellcolor{gray!20}$\textit{\underline{92.64}}_{\pm0.44}$ & \cellcolor{gray!20}$\textbf{93.30}_{\pm0.37}$ & \cellcolor{gray!20}$\textbf{93.68}_{\pm0.52 }$ & \textcolor{blue}{\cellcolor{gray!20}$\textbf{93.86}_{\pm0.36}$}  \\
\texttt{AGSSL-TS}             & \cellcolor{gray!20}$72.72_{\pm0.36}$ & \cellcolor{gray!20}$\textit{\underline{73.24}}_{\pm0.44}$ & \cellcolor{gray!20}$\textit{\underline{73.34}}_{\pm0.40}$ & \cellcolor{gray!20}$\textit{\underline{73.58}}_{\pm0.38 }$ & \textcolor{blue}{\cellcolor{gray!20}$\textit{\underline{74.20}}_{\pm0.42 }$} & \cellcolor{gray!20}$92.30_{\pm0.67}$ & \cellcolor{gray!20}$\textbf{92.88}_{\pm0.34}$ & \cellcolor{gray!20}$\textit{\underline{93.04}}_{\pm0.26}$ & \cellcolor{gray!20}$\textit{\underline{93.38}}_{\pm0.31 }$ & \textcolor{blue}{\cellcolor{gray!20}$\textit{\underline{93.46}}_{\pm0.25}$}  \\ \bottomrule

\end{tabular}} \vspace{-1em}
\end{center}
\end{table*}

\subsection*{I. Details on Experimental Results}
Table.~\ref{tab:A2} provides the concrete values of experimental results in Fig.~\ref{fig:8a} and Fig.~\ref{fig:8b}. The setting of different teacher combination is (1) one teacher: \texttt{PAR}; (2) two teachers: \texttt{PAR} and \texttt{CLU}; (3) three teachers: \texttt{PAR}, \texttt{CLU}, and \texttt{DGI}; (4) four teachers: \texttt{PAR}, \texttt{CLU}, \texttt{DGI}, and \texttt{PAIRDIS}; and (5) five teachers: \texttt{PAR}, \texttt{CLU}, \texttt{DGI}, \texttt{PAIRDIS}, and \texttt{PAIRSIM}. As can be seen from Table.~\ref{tab:A2}, \texttt{AGSSL-LF} and \texttt{AGSSL-TS} always perform better than other three heuristic methods; their performance increases consistently with the number of teachers, reaching the best at a number of five teachers.

\begin{table*}[!htbp]
\begin{center}
\caption{Comparison of the computational resources in terms of training time (\texttt{s}) and peak memory (\textit{M}) among \texttt{AGSSL-LF}, \texttt{AGSSL-TS} and the \textit{Joint Training} (\texttt{JOINT-T}) on eight graph datasets.}
\vspace{0.5em}
\label{tab:A3}
\resizebox{\textwidth}{!}{
\begin{tabular}{llcccccccccc}

\toprule
\textbf{Method} & \textbf{Resource} & \texttt{Cora} & \texttt{Citeseer} & \texttt{Pubmed} & \texttt{CS} & \texttt{Physics} & \texttt{Photo} & \texttt{Computers} & \texttt{ogbn-arxiv} \\ \midrule
\multirow{2}{*}{\texttt{JOINT-T}}    & Training Time    & 17.87s & 18.57s & 75.18s & 98.83s & 171.61s & 36.73s & 51.90s & 1362.28s \\
                            & Peak Memory & 1527M  & 1915M  & 7415M  & 10235M & 22851M  & 4358M  & 4478M  & 24573M   \\ \midrule
\multirow{2}{*}{\texttt{AGSSL-LF}} & Training Time    & 15.42s & 16.09s & 68.31s & 91.76s & 158.73s & 32.61s & 45.96s & 1289.73s \\
                            & Peak Memory & 1422M  & 1618M  & 5834M  & 6512M  & 14395M  & 3693M  & 3847M  & 15739M   \\ \midrule
\multirow{2}{*}{\texttt{AGSSL-TS}} & Training Time    & 15.53s & 16.22s & 68.67s & 92.14s & 159.24s & 32.84s & 46.26s & 1294.67s \\
                            & Peak Memory & 1453M  & 1695M  & 5912M  & 6624M  & 14538M  & 3752M  & 3915M  & 15923M   \\ \bottomrule

\end{tabular}}
\end{center}
\end{table*}

\subsection*{J. Computational Efficiency}
We compare the training time and peak memory of \texttt{AGSSL} with the joint training (\texttt{JOINT-T}) of multiple pretext tasks with fixed weights in Table.~\ref{tab:A3}. It can be seen that while \texttt{AGSSL} needs to train multiple teacher models separately, it has advantages over \texttt{JOINT-T} in terms of both training time and peak memory, mainly because: (1) compared to \texttt{JOINT-T} that trains multiple pretext tasks \textbf{simultaneously}, each teacher in \texttt{AGSSL} can be trained \textbf{sequentially}, which greatly reduces the peak memory usage; (2) the training with multiple tasks is more difficult to optimize than the training with one single task, so \textbf{each training epoch of \texttt{JOINT-T} takes longer time} than \texttt{AGSSL}; and (3) \texttt{JOINT-T} is more difficult to converge with higher complexity, i.e., \textbf{it requires more training epochs to converge}. Instead, while \texttt{AGSSL} needs to train several more models, it takes much less time for each model, resulting in an overall training time even slightly less than \texttt{JOINT-T}. (4) \texttt{AGSSL-LF} and \texttt{AGSSL-TS} differ only in their knowledge integration schemes, so their overall resource usage (e.g., training time and peak memory) is very close and much less than \texttt{JOINT-T}.

\clearpage

\end{document}